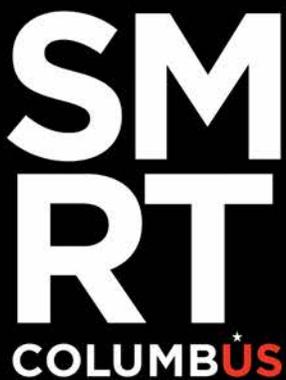

# Simulation Environment for Safety Assessment of CEAV Deployment in Linden

for the Smart Columbus Demonstration Program

**FINAL REPORT | November 7, 2019**

## Notice

## Acknowledgement of Support

This material is based upon work supported by the U.S. Department of Transportation under Agreement No. DTFH6116H00013.

## Disclaimer

# Acknowledgements

The Smart Columbus Program would like to thank the following authors at The Ohio State University for their support of Smart Columbus and their contributions to this report that supports the Connected Electric Autonomous Vehicles deployment.

| | |
|---|---|
| Levent Guvenc, The Ohio State University | Bilin Aksun-Guvenc, The Ohio State University |
| Xinchen Li, The Ohio State University | Sheng Zhu, The Ohio State University |
| Karina M. Meneses Cime, The Ohio State University | Aravind Chandradoss Arul Doss, The Ohio State University |
| Sukru Yaren Gelbal, The Ohio State University | Automated Driving Lab, The Ohio State University |

The Smart Columbus Program would also like to thank the reviewers of this report.

| | |
|---|---|
| Andrew Wolpert, City of Columbus | Jeff Kupko, Michael Baker International |
| Tom Timcho, WSP | |

# Abstract

This report presents a simulation environment for pre-deployment testing of the autonomous shuttles that will operate in the Linden Residential Area. An autonomous shuttle deployment was already successfully launched and operated in the city of Columbus and ended recently. This report focuses on the second autonomous shuttle deployment planned to start in December, 2019, using a route that will help to solve "first-mile / last-mile" mobility challenges in the Linden neighborhood of Columbus by providing free rides between St. Stephens Community House, Douglas Community Recreation Center, Rosewind Resident Council and Linden Transit Center. This document presents simulation testing environments in two open source simulators and a commercial simulator for this residential area route and how they can be used for model-in-the-loop and hardware-in-the-loop simulation testing of autonomous shuttle operation before the actual deployment.

# Table of Contents

# Executive Summary

This document focuses on the development of a simulation environment for pre-deployment testing of the autonomous shuttles that will operate in the Linden area. Autonomous shuttles that emulate SAE Level L4 automated driving using human driver assisted autonomy have been operating in geo-fenced areas in several cities in the US and the rest of the world. As the winner of the U.S. Department of Transportation's (USDOT) first-ever Smart City Challenge, Columbus was awarded $40 million in grant funding and the designation as America's Smart City. The autonomous shuttle deployment in the Linden area is one of the deployments of the Smart Columbus project aimed at improving the mobility choices of residents of Columbus in a transportation challenged part of the city.

An autonomous shuttle deployment was already successfully launched and operated in the city of Columbus and ended recently. This deployment was called the Smart Circuit and circled the Scioto Mile in downtown Columbus, providing free rides to the Center of Science and Industry (COSI), the National Veterans Memorial and Museum, Bicentennial Park and the Smart Columbus Experience Center. This document focuses on the second autonomous shuttle deployment planned to start in December, 2019, using a route that will help to solve "first-mile / last-mile" mobility challenges in the Linden neighborhood of Columbus by providing free rides between St. Stephens Community House, Douglas Community Recreation Center, Rosewind Resident Council and Linden Transit Center.

Autonomous shuttle operation in urban areas populated by pedestrians should be thoroughly tested before deployment. This document mainly presents a simulation testing environment in an open source simulator and how it can be used for model-in-the-loop and hardware-in-the-loop testing of autonomous shuttle operation before the actual deployment. Other approaches to creating the simulation environment and simulation testing of autonomous driving and how to add traffic are also presented using another open source simulator and a commercially available simulator. The document also reports autonomous driving sensor data collection in the Linden route.

The organization of the report is as follows. Data collection along the autonomous shuttle route in Linden is presented first in Chapter 1. Simulation environments based on a real world map are constructed and presented in Chapter 2. The characteristics of the two open source simulators are presented in Chapter 3 in the context of the Linden Residential Area route. Use of the simulation environment created and the simulators for testing autonomous driving is presented in Chapter 4. A commercially available simulator is used in Chapter 5 for the same autonomous driving simulation testing where an NVIDIA Drive PX2 is used to run autonomous driving functions like free space determination, object detection and classification and lane detection. How the simulation environment and simulator can be used to test the autonomous driving system under different test cases and in the presence of other traffic is treated in Chapter 6. The report concludes with significant events and traffic situations that were identified during simulations, the problems they may cause and possible mitigation methods in Chapter 7.

# Chapter 1. Data Collection in the Linden Residential Area and the Scioto Mile Smart Circuit Routes

Autonomous vehicles combine a variety of sensors to localize themselves and perceive their surroundings, including Radio Detection and Ranging (radar); Light Imaging, Detection and Ranging (lidar), camera, Sound Navigation and Ranging (sonar), Inertial Measurement Unit (IMU) and Global Positioning System (GPS) sensors among others. The information provided by these sensors is used to identity navigation paths, avoid obstacles and read relevant markers, like road signs. In multiple locations around the world, autonomous vehicle development teams run tests that result in thousands of hours of test drive data. One eight-hour data collection drive can create more than 100 terabytes of data. This massive amount of data must be collected, offloaded, stored, and interpreted for algorithmic training to build perception and decision-making algorithms. Lidar and/or camera data is used to build a map of the routes taken if a map matching approach to localization is going to be used. Lidar and/or camera data may also be used to build a realistic simulation environment.

The sensors used in the unified connected and autonomous driving architecture [1] developed by the Automated Driving Lab (ADL) at the Ohio State University (OSU) and the associated sensor library were used to collect and record autonomous driving sensor data along the Scioto Mile Smart Circuit and Linden Residential Area autonomous shuttle routes.

## 1.1. UNIFIED ARCHITECTURE AND LIBRARY

In this section, the unified architecture sensor library used for data collection is presented. The vehicle used for the data collection is the Ohio State University Automated Driving Lab research level Autonomous Vehicle (AV) shown in Figure 1. It is a 2017 Ford Fusion hybrid vehicle with full drive-by-wire capability, computers and electronic control unit for automated driving functions and a range of perception, localization and communication sensors. The sensors and computer(s) relevant to the autonomous driving data collection of this chapter are also shown in Figure 1.

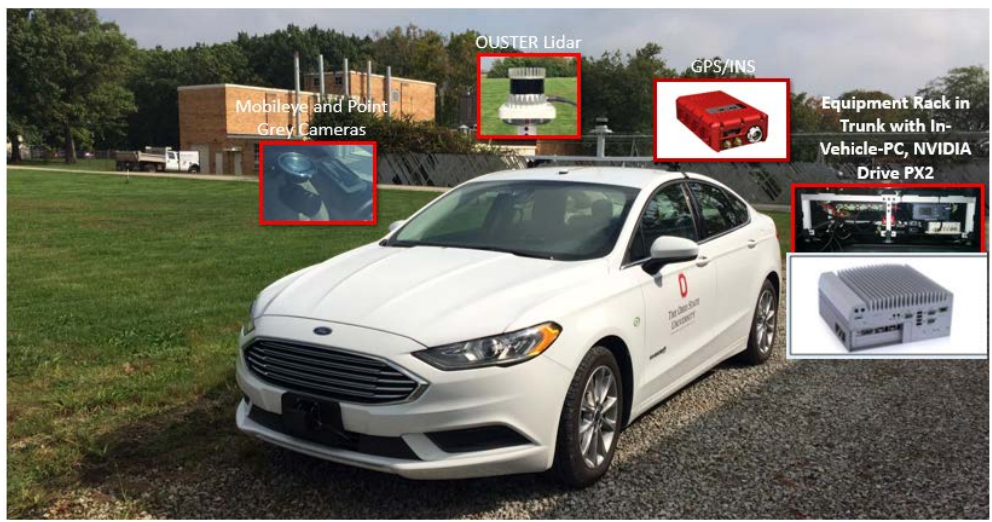

**Figure 1: Autonomous Driving Data Collection Sensors in the OSU ADL Research AV**

As seen in Figure 2, the autonomous driving sensors in the OSU ADL AV include forward looking cameras alongside with Mobileye intelligent camera, 3D lidar, radars that are forward looking and side looking on 4 corners, high accuracy GPS unit with RTK bridge for GNSS localization and DSRC OBU for connected vehicle application for V2V and V2I communication. In the data collection reported here, the forward looking camera with a maximum of 30 fps recoding rate along with the 3D 64-channel high definition lidar provide the ability of high resolution environment sensing and object/landmark detection. The high accuracy GPS unit used is packaged with IMU as well as an RTK bridge for over the air NTRIP corrections for improved localization accuracy during the data collection.

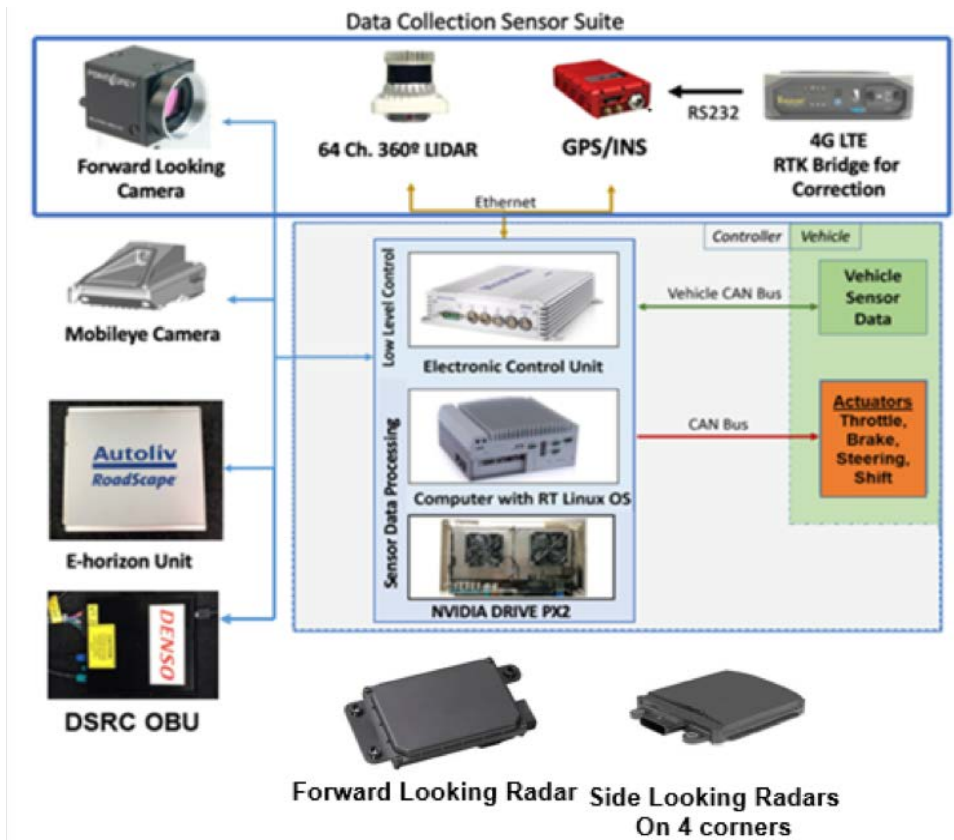

**Figure 2. OSU ADL AV Hardware Architecture with Sensors Used**

## 1.2. DATA COLLECTION

We used the OSU ADL AV in Figure 1 to collect multiple sensor raw data for use in autonomous driving research and to help autonomous shuttle operators who will deploy their vehicles in those routes. The Scioto Mile Smart Circuit route where autonomous shuttles by May Mobility operated as a circulator until recently is displayed in Figure 3. Data was first collected on this route as it was the only operational AV shuttle route in Columbus at the time. The OSU Automated Driving Lab also built a 3D simulation model, updated the OpenStreetMaps (OSM) [2] map and prepared a Vissim [3] traffic simulation model for this route. These results are not presented in this report as the focus is on the Linden Residential Area AV shuttle deployment that will start in December 2019. The Linden Residential Area autonomous shuttle circulator route is shown in Figure 4. The experience gained in autonomous driving data collection was used in the data collection drive(s) for the Linden Residential Area route, resulting in a faster and more efficient data collection and storage exercise. Our experience with the Scioto Mile Smart Circuit route also allowed us to build a lidar point cloud based map faster and more efficiently. AV shuttle operators also map the route first and build a similar map which is later

4 | Smart Columbus Program | Linden Residential Area CEAV Simulation Evaluation – Final Report

used for map matching based localization which can also work in the presence of low GPS positioning accuracy or loss of GPS signals.

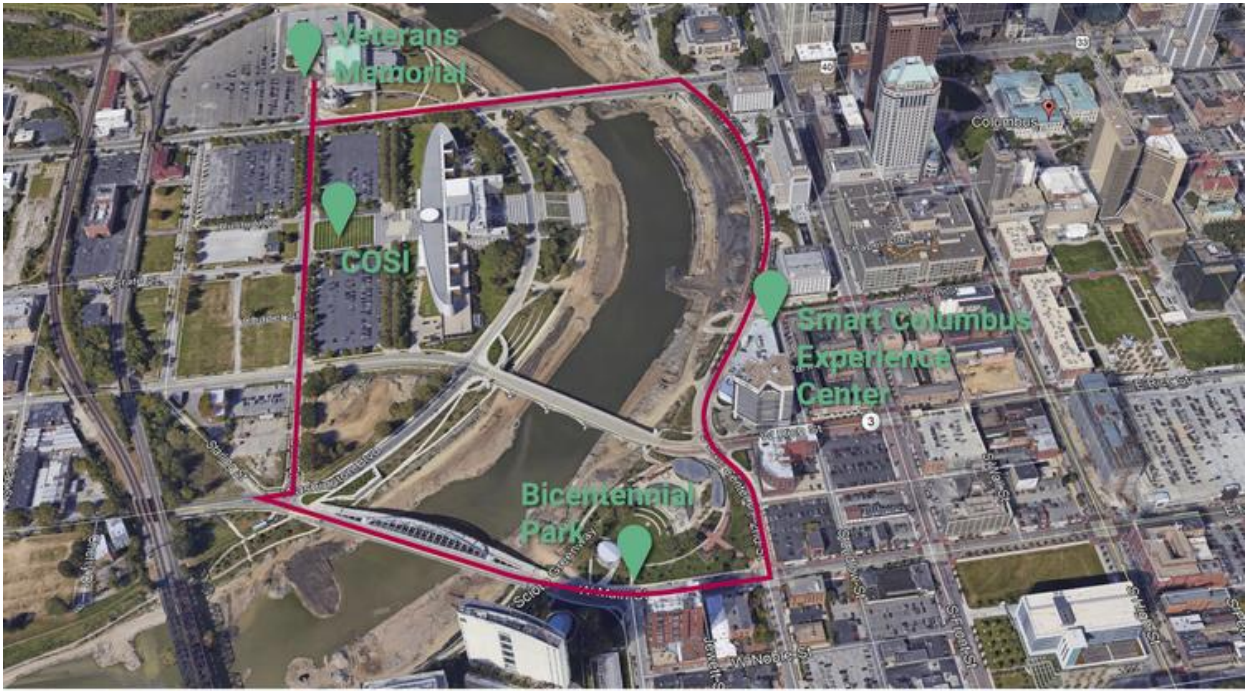

**Figure 3. Scioto Mile Smart Circuit AV Shuttle Route**

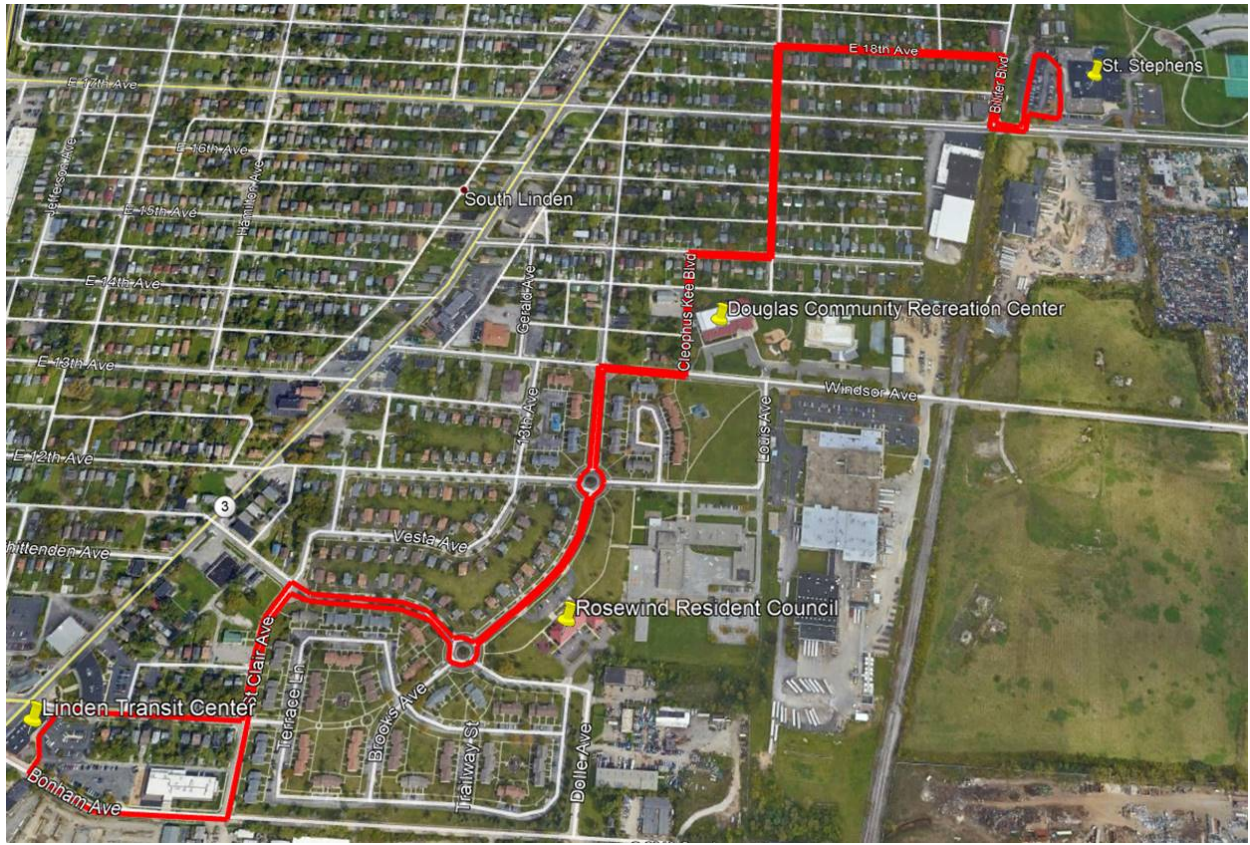

**Figure 4. Linden Residential Area AV Shuttle Route**

The data collection in the Scioto Mile Smart Circuit and Linden Residential Area routes has resulted in different groups of data. The collected data has been shared using a publicly available link to a box site of OSU. Due to limitation of file size in this box site, the data from a single data collection drive is separated into smaller sized files. The box site also contains information on the file contents and which sensor data was recorded. The recorded data is classified based on the data collection area and the sensors used for data collection. Further development will include data from other areas and data labelling for various ways of utilization. The link for accessing the collected data is given as: https://osu.box.com/s/ygzh1t7a6pgn7ngki62k7kjxw20q60kn. Screenshots of visualization of data collected on the Scioto Mile Smart Circuit and the Linden Residential Area routes are presented in Figures 5 and 6, respectively.

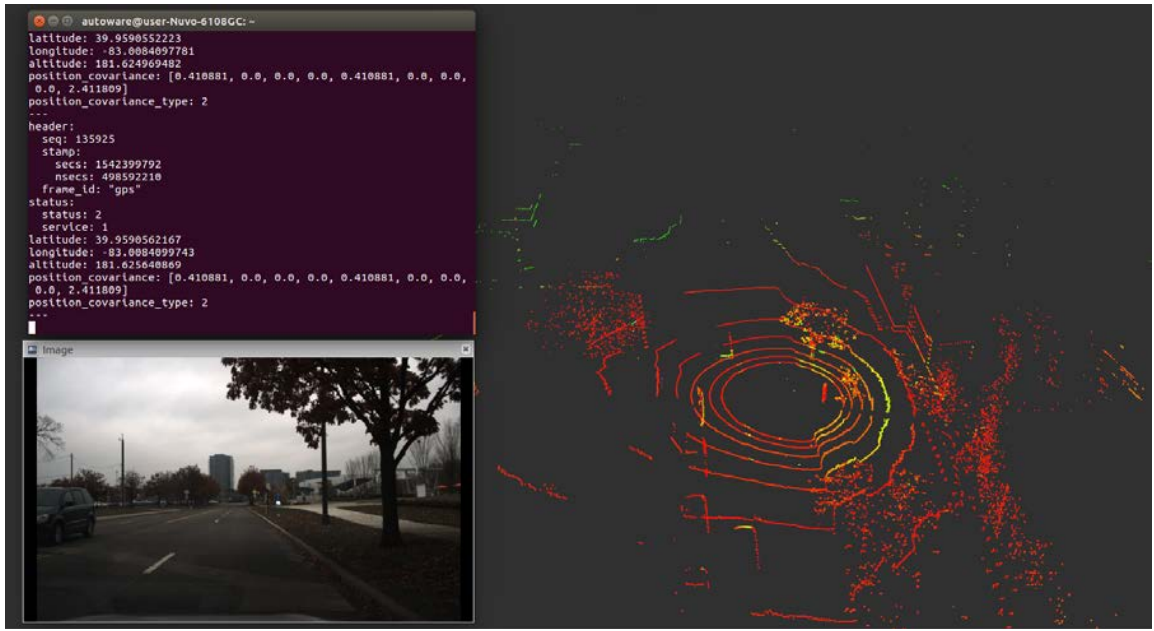

**Figure 5. Screenshot of AV Data Collected in Scioto Mile Smart Circuit Route**

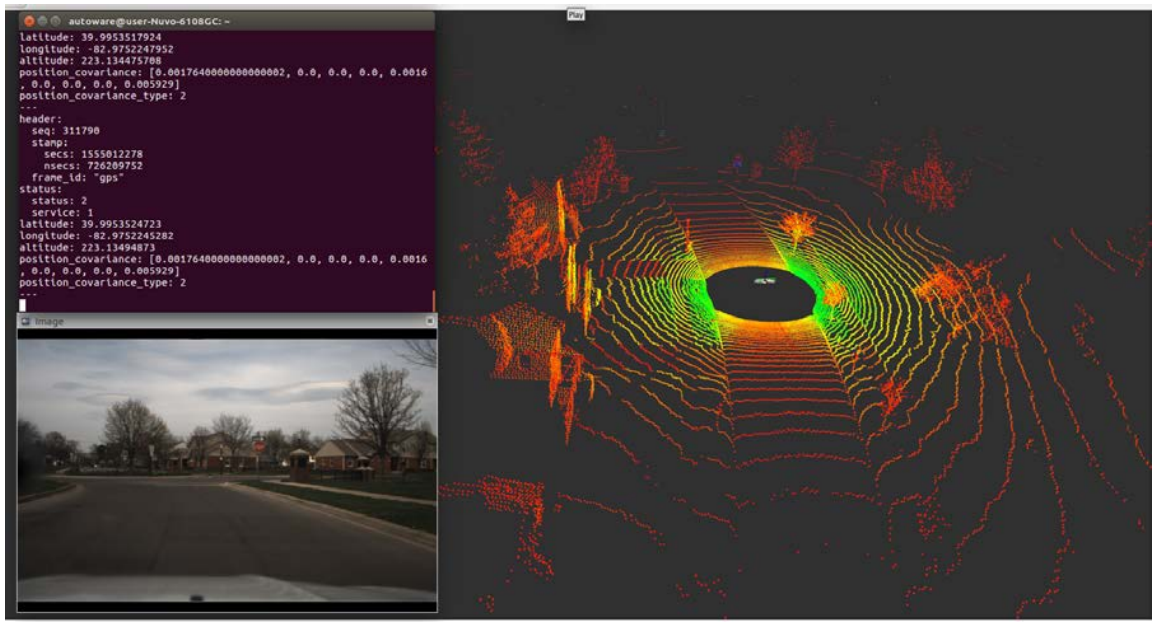

**Figure 6. Screenshot of AV Data Collected in Linden Residential Area AV Shuttle Route**

# Chapter 2. Simulation Environment Construction Based on Real-World Data

For Model-in-the-Loop (MiL) and Hardware-in-the-Loop (HiL) simulation and validation of autonomous driving functions and safety evaluation of a planned deployment using simulation tools, a realistic virtual environment is extremely important from many aspects, including use of a correct map of roads, positions of objects in the environment, road infrastructure and features in the environment and the ability to add traffic and Vulnerable Road Users (VRU) like pedestrians and bicyclists.

In our work, different open source platforms for building the simulation environment and running the autonomous driving simulations were studied and evaluated for simulation testing of low speed autonomous shuttles in pre-determined urban routes in geo-fenced areas. The process of creation and construction of the simulation environment for the Linden Residential Area on different platforms is introduced in this chapter.
To ensure the accuracy of the map and repeatability of results and scalability to other deployment locations, the open source and freely available OpenStreetMap map was used for creating the simulation environment road geometry. For creating objects in the environment of the map and road infrastructures, multiple rendering algorithms were used for creating the corresponding meshes and object surfaces. The executable file of the simulator and the source of simulator is placed in a shared box folder at the link of https://osu.box.com/s/7z3sgqkuj2grtclouybxwykproaxg013, anyone who has the link will have access to download the simulator with Linden area and a test scene of the city of San Francisco.

## 2.1. BUILDING THE ROAD NETWORK AND MAP

Currently, many maps are publicly available for everyday use such as Google Maps or Bing maps. High definition maps that are specifically built for autonomous driving are also available for highways from companies like HERE and TomTom. Different from those commercial maps, the OpenStreetMap (OSM) is a freely available, open source map. Thus, the users have access to the geodata underlying the map. The data from OSM can be used in various ways, including production of paper maps and electronic maps (similar to Google Maps, for example), geocoding of address and place names, and route planning. As it is open source, it is editable through different scripts. The road network in OSM map is relatively complete so that it can be used in navigation and traffic simulations. Since OSM is based on user contributed data, there are also problems that need to be fixed before using it to create the road network in a simulator. The problems include possibly incorrect number of lanes, possibly incorrect intersection geometry and speed limits not being up to date. Before using OSM, the map has to be updated and shared with the rest of the user community. In rural, residential areas, significant objects like residential housing, buildings and some road signs are not modelled and mapped. This, of course, is due to the map not being up to date and causes a lack of environmental information which affects the realism of creating the autonomous driving simulation environment based on the OSM map. Thus, using scripts such as the Java OpenStreetMap (JOSM) editor, the OSM map information was improved by updating incorrect or missing information and adding complete housing information along the Linden Residential Area so that it can be used in our simulation environment development as shown in Figure 7. This required extensive manual work but the updated Linden Residential Area OSM map has been pushed to the OSM server to update the source map and is now available to the whole user community of OSM maps.

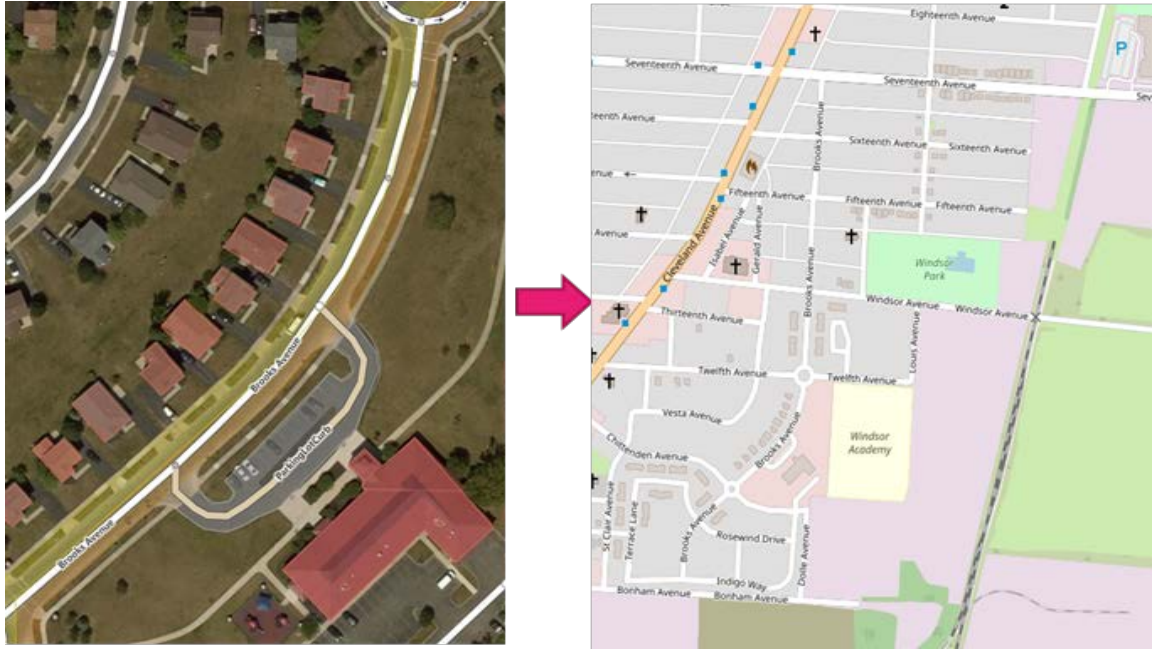

**Figure 7. Editing OpenStreetMap and Exporting It for Use in Simulation**

## 2.2. SIMULATION ENVIRONMENT CONSTRUCTION ON DIFFERENT PLATFORMS

### 2.2.1. Environment Construction on the Unity Platform

The Unity platform is a cross-platform game engine developed by Unity Technologies for real-time, three dimensional graphics visualization. The engine has been adopted in many aspects outside the field of video games such as filming, automotive CAD, architecture and construction. With the rise of the autonomous vehicle industry, the Unity platform is now applied more and more in the field of autonomous driving. The Unity platform is one of the platforms that can be used for visualization of the autonomous driving environment and for extracting autonomous vehicle raw data. The freely available, open source LG SVL simulator [4] which is the main autonomous shuttle simulator in this report for emulating autonomous shuttle operation in the Linden Residential Area is based on the Unity platform.

Various Unity scripts and assets were used for importing the OSM environment. The imported map has complete environment information containing the road network, traffic infrastructures, housing and other related objects. A pipelined approach was used as shown in Figure 8.

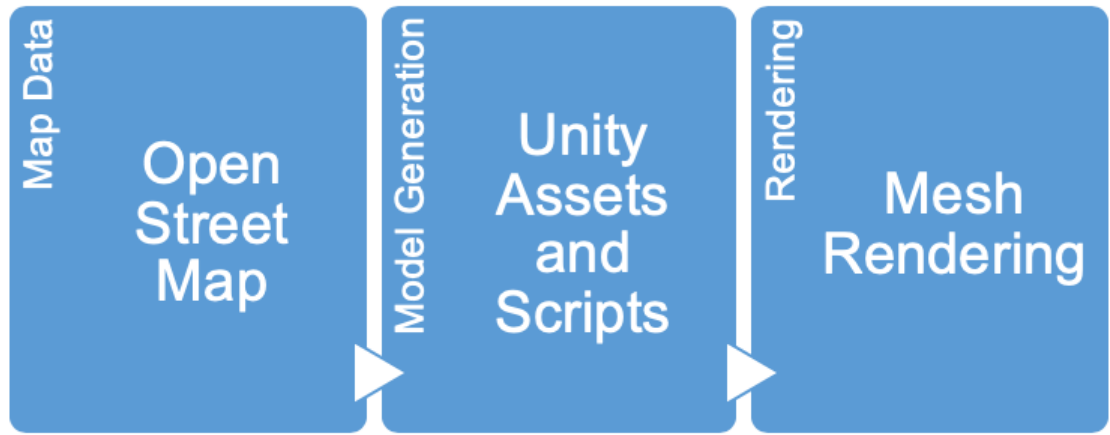

**Figure 8. Pipeline for Importing OSM Map into Unity**

After editing the OSM map according to available satellite images and height images, the map was imported into the Unity Engine. At this stage, the map in Unity contains the road network and the raw mesh without any texture which looks different from the corresponding real world appearance as shown in Figure 9. Afterwards, road signs and road features were also added and rendered in the Unity Engine so that the real world environment can be completely emulated in the simulations. Those objects were rendered manually by comparing with other objects in the simulation environment.

Different methods for rendering the meshes were used to better represent the real world objects as shown in Figure 10. Different mesh generating methods were tried and used for mesh generation and rendering, for example MeshLab [5], *asset* in the Unity Asset Store and so on. The Easy Road 3D Asset Package [6] in Unity platform was used to render the road surface. Default meshes available in the Unity project for surface rendering were used for the buildings.

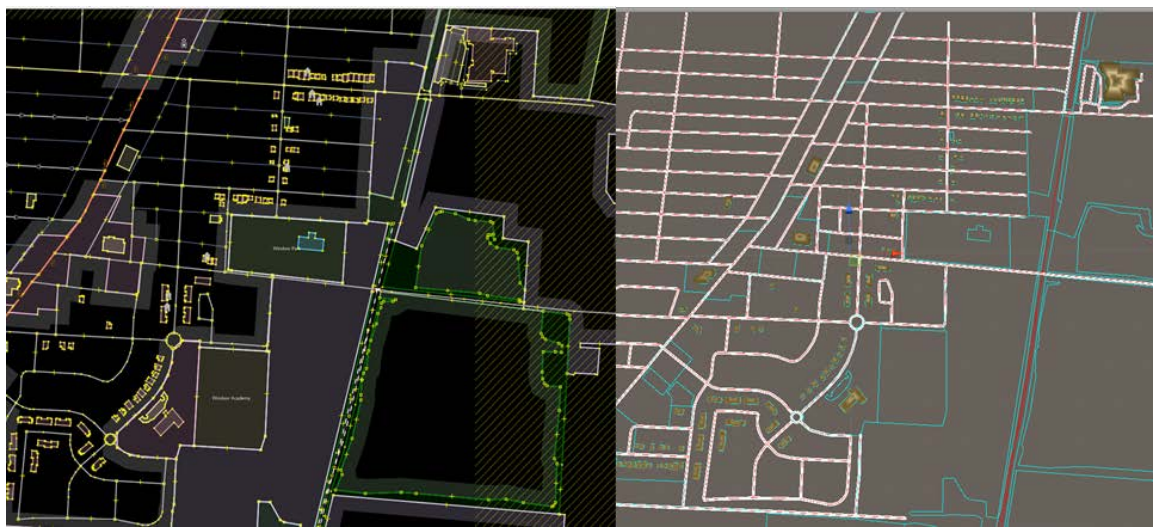

**Figure 9. OpenStreetMap in Linden Residential Area (Left) and Corresponding Un-edited, Imported Map in Unity**

## 2.2.2. Environment Construction on the Unreal Engine Platform

This section presents simulation environment development based on the Unreal Engine platform. The freely available and open source CARLA simulator [7] is used for the Unreal Engine based simulations. The pipelined approach for creating the Unreal Engine simulation environment as shown in Figure 11 is introduced here in the context of the Linden Residential Area route. The overall aim is to use time-stamped three dimensional (3D) lidar data to reconstruct the surface of the entire environment and then to add texture details from Google Maps (Google Earth Studio [8]) and road details from OpenStreetMaps.

**Pre-processing**. In this step, we mainly focus on reducing the redundant point cloud data to boost the processing speed without considerable loss in reconstruction. Due to lidar construction, point cloud density is dependent on measurement distance, i.e., the point cloud will be very dense near the AV and sparse at further distance. Therefore, we use the Poisson Sampling Disk approach to remove highly cluttered point clouds using sampling disks. By using this step, it is possible to reduce up to 40-45% of lidar point cloud data. The recorded lidar data usually contains point clouds corresponding to other vehicles on the road. It is highly recommended to remove any such unwanted point clouds that correspond to objects that are stationary for a significant time during recording, for example, any stationary car waiting for a traffic signal. One can remove such point clouds in the pre-processing step using the SqueezeSeg [9] Convolutional Neural Network (CNN) or other start-of-art methods. During the pre-processing step for the Linden Residential Area recorded data, the freely available and open source MeshLab, and Point Cloud Library (PCL) [10] tools were used.

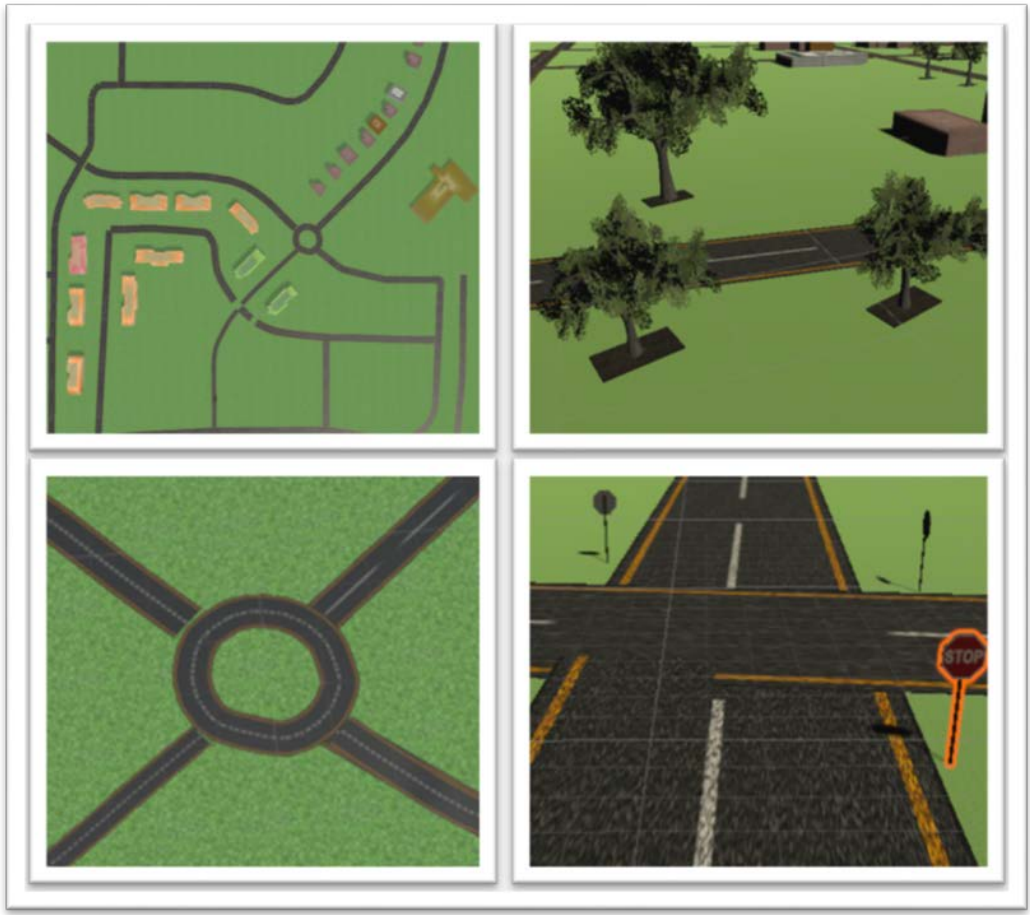

**Figure 10. Mesh Generation and Rendering in Unity**

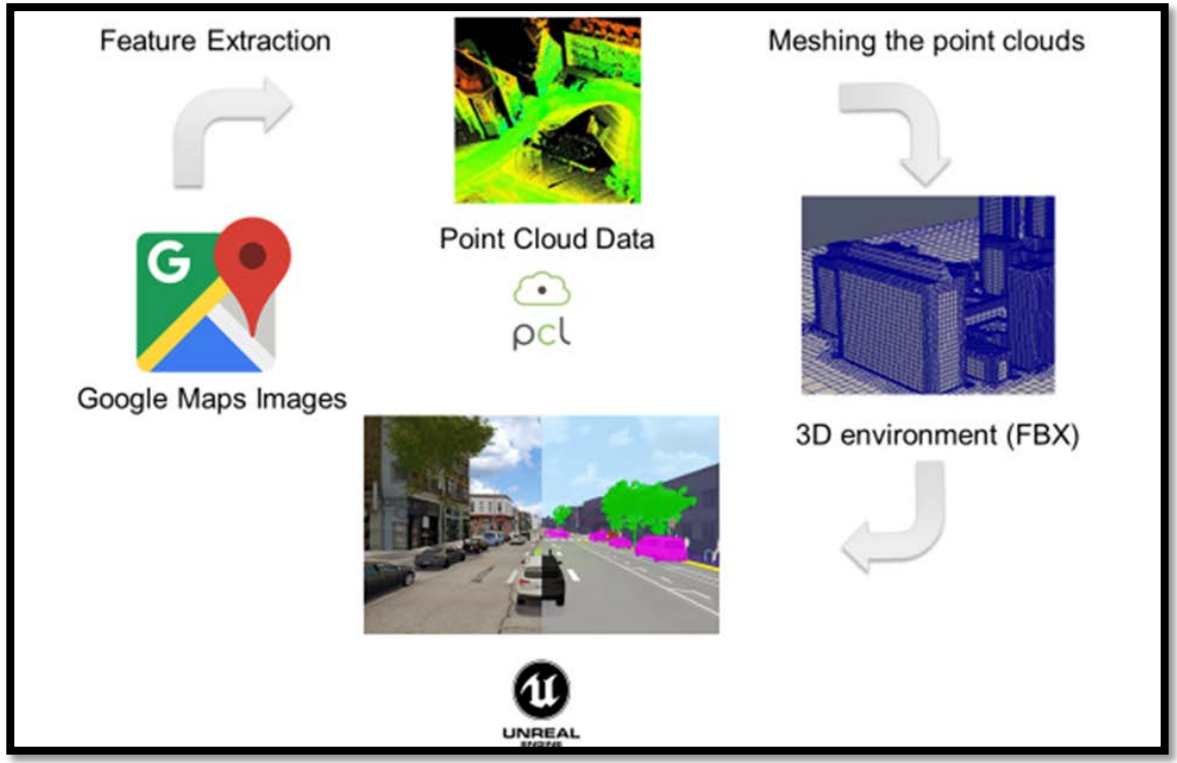

**Figure 11. Pipelined Approach Used for Importing Real Maps into Unreal Engine**

**Mesh Generation**. In this step, a mesh is created from the point cloud data and is smoothed using a smoothening filter after which textures is added. Vertex normals to the point cloud are calculated first. The parameters for calculating the normals are selected heuristically based on lidar resolution, size of simulation environment and other similar factors. With the generated normals, the surface is constructed by building a mesh using the point cloud data. The Ball-Pivoting surface reconstruction was seen to give the best results. The parameters for Ball-Pivoting were selected based on point cloud density, size of simulation environment and others factor. In practice, surface reconstruction results in a mesh with voluminous triangles. Quadratic edge decimation was used to reduce the triangle count. Faulty triangles including duplicate edges and single point triangles were removed using PCL and MeshLab tools. A Laplacian filter was used next to smoothen the surface. Texture from Google Maps is added to the mesh in the next step to make the simulation environment look photo-realistic for simulation. The freely available and open source MeshLab, Blender [11], and PCL tools were used for mesh generation for the Linden Residential Area.

Nearly 4,000 satellite images of the Linden Residential Area route were retrieved from Google Earth Studio [7] and used for generating mesh texture. It is also possible to collect data using drones in order to construct a more realistic and up to date replication of the environment. It is not possible to recover texture information in occluded regions such as those under the trees.

In the upcoming steps, only the important tasks required to achieve the desired results are presented while skipping intermediary steps like feature extraction, point-to-point correspondence and feature matching which are easily taken care of using tools from the freely available, open source Meshroom [12] program.

**Structure from Motions (SfM) and Depth Map**. Using existing photogrammetry tools, with the matched features from previous step, the 3D reconstruction of the environment was achieved using SfM. Meshroom's SfM implementation was used here. The parameters for SfM were selected heuristically. Similar results can also be achieved using the freely available, open sourced VisualSFM tool. The mesh for the Linden Residential Area route was generated from 3D lidar scan data and the texture was only added for a photo realistic look. A depth map can also be generated using the dense point cloud from Google Earth images in order to obtain a better reconstruction of the environment.

**Texture Generation**. The surface generated from the meshing step is combined with the texture from the dense SfM using Least Square Conformal Maps. Blender was used in order to change the origin before using the mesh for texture generation. Once the textures are added to the surface, Blender and PCL libraries were used to convert the file format to filmbox (.fbx) as it compatible with both the Unreal Engine and Unity platforms.

The resulting rendered mesh with texture for the Linden Residential Area AV route is shown in Figure 12. Figure 13 displays close up views of select parts of that meshed simulation environment.

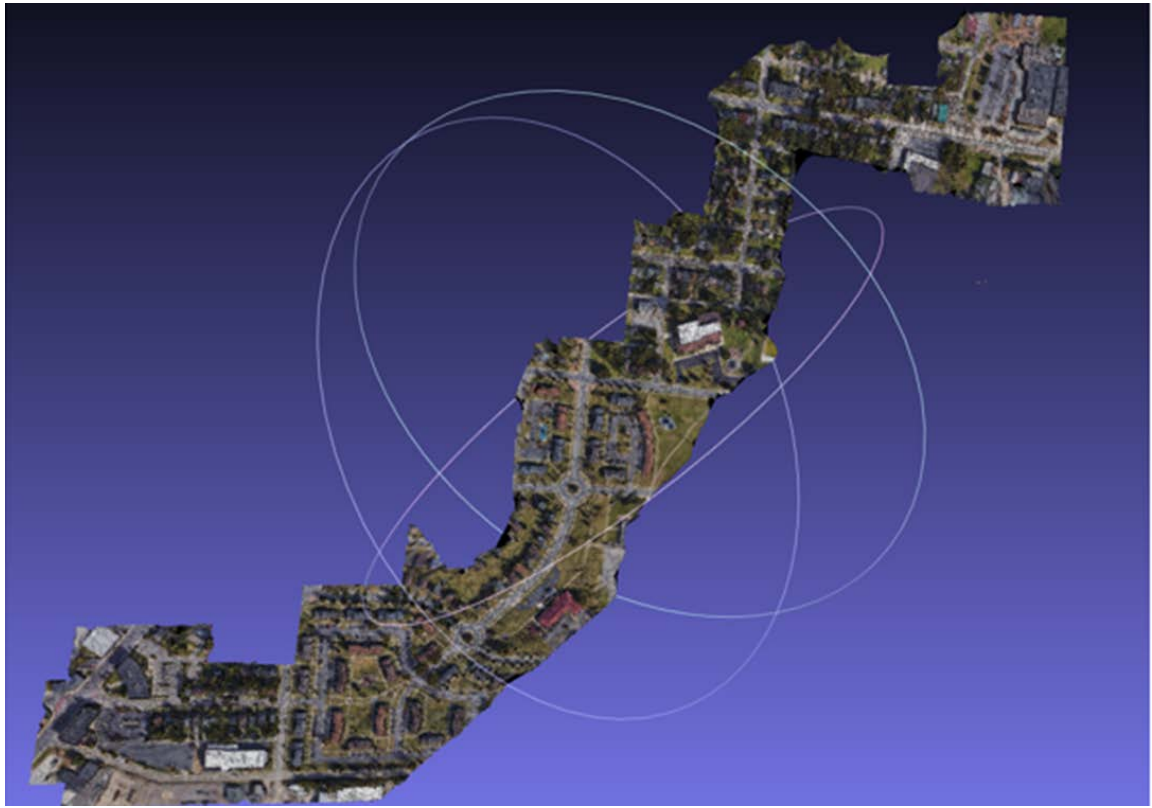

**Figure 12. Linden Residential Area AV Route as a Rendered Mesh**

### 2.2.3. Environment Construction in CarMaker

The two simulation environment generation approaches presented earlier in this chapter were based on the use of freely available and open source code and platforms. This section introduces simulation environment building in a commercially available simulator. CarMaker is used here as the specific commercial simulator software. As in the previous two approaches, OSM map data is used again to create a realistic road network in CarMaker as shown in Figure 14. The Vissim interface of CarMaker is used to model microscopic traffic and realistically incorporate it into the AV shuttle simulations.

14 | Smart Columbus Program | Linden Residential Area CEAV Simulation Evaluation – Final Report

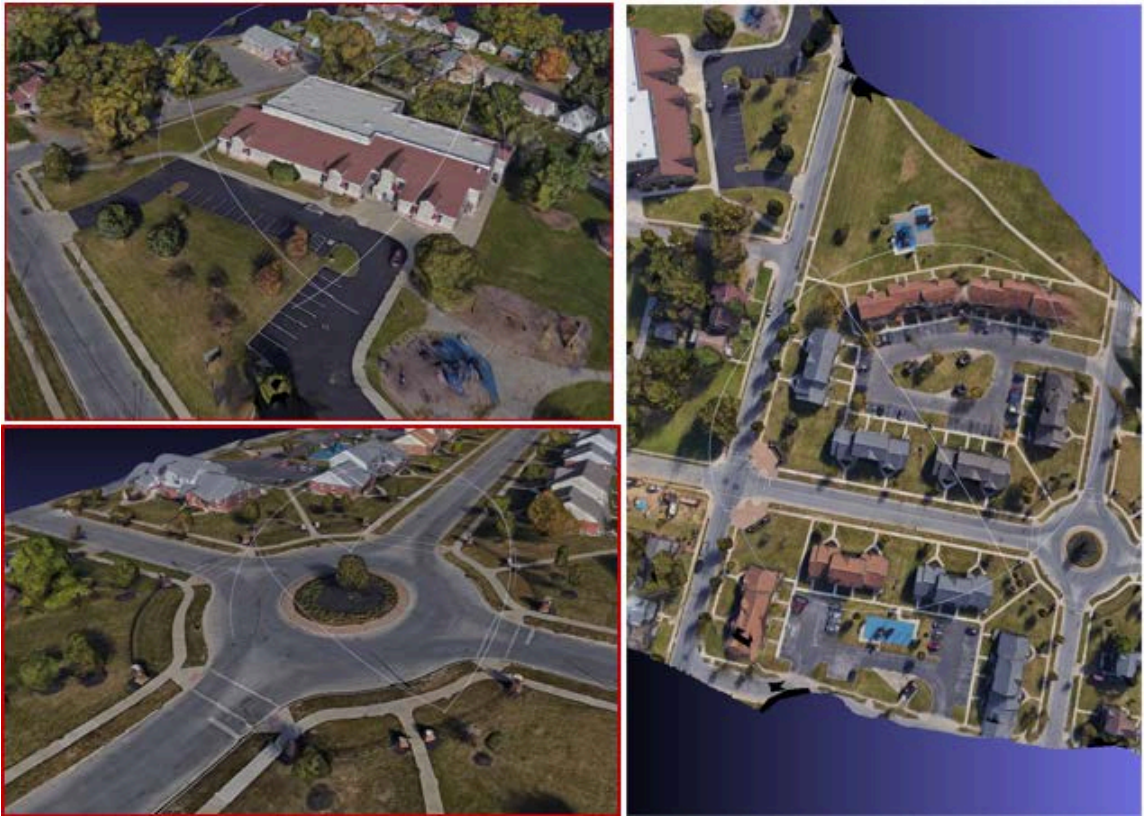

**Figure 13. Local Features of the Mesh in the CARLA Simulator**

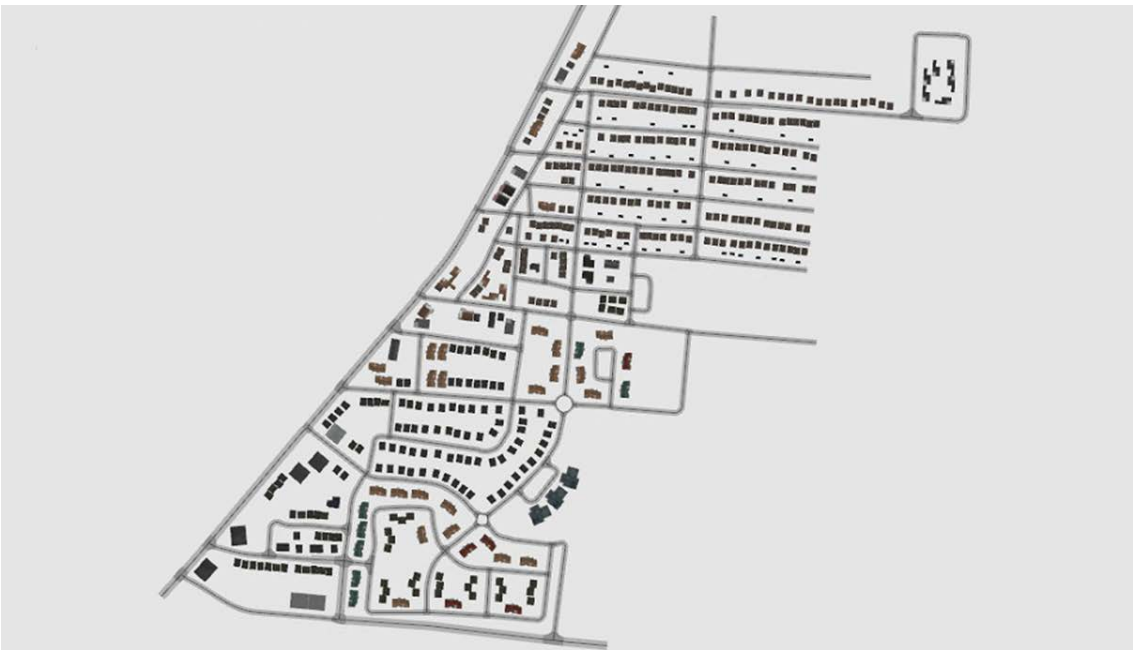

**Figure 14. Map Creation in CarMaker**

Since CarMaker does not import other details like buildings and traffic infrastructure directly form a map, these were entered manually. Figure 15 shows the Linden Residential Area simulation environment in CarMaker where buildings and trees were added for a more realistic view.

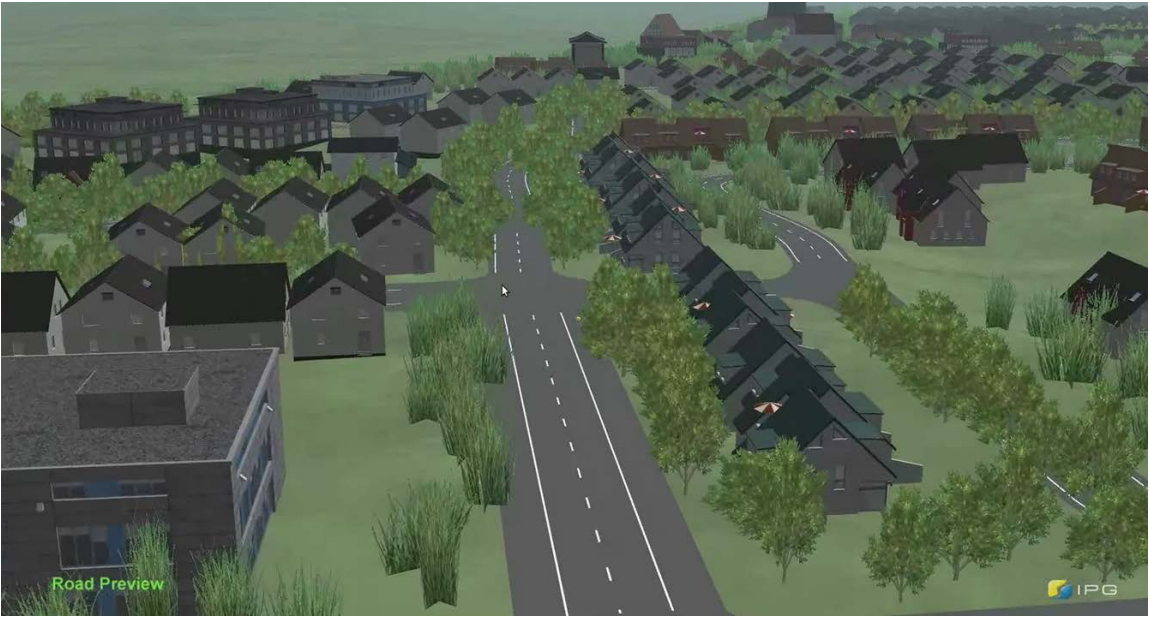

**Figure 15. Linden Residential Area Simulation Environment in CarMaker**

# Chapter 3. Characteristics of the Open Source Simulation Environments and Simulators Used

## 3.1. UNITY BASED SIMULATION

The Unity based simulation environment developed here for the Linden Residential Area is used in the freely available and open source LG SVL simulator. The virtual sensors used in the LG SVL simulator are three dimensional (3D) lidar, cameras, radar, GPS and IMU of the AV in the simulation. The 3D lidar point cloud emulation of the LG SVL simulation in the Unity platform is shown in Figure 16. During the simulation with a computer with Central Processing Unit (CPU) and Graphical Processing Unit (GPU), most of the ray casting based lidar computations are handled by the GPU while the distance information is calculated in the CPU. The emulation of 3D lidar point cloud data generation in the Linden Residential Area Unity based LG SVL simulation is shown in Figure 17.

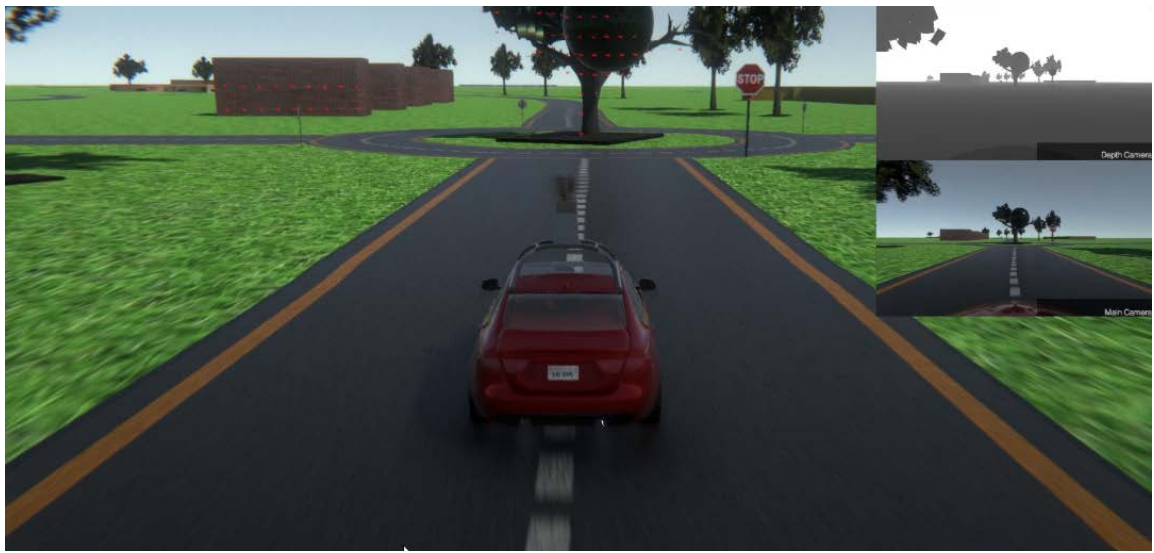

Figure 16. Unity Based LG SVL Simulatior with 3D Lidar and Camera

The Unity based LG SVL simulation camera renders pixels within the preset range and provides Red-Green-Blue (RGB) color and depth information per pixel in the camera data which is illustrated in Figure 18. Besides the functionality of emulating essential AV sensors, the Unity Engine based simulation environment has the capability of simulating Non-Player Characters (NPC) which are used to model the other vehicles on the road. These NPC vehicles follow pre-defined lanes and obey traffic rules. It is possible to investigate the operation of the AV within traffic in the LG SVL simulator using the NPC vehicles as shown in Figure 19.

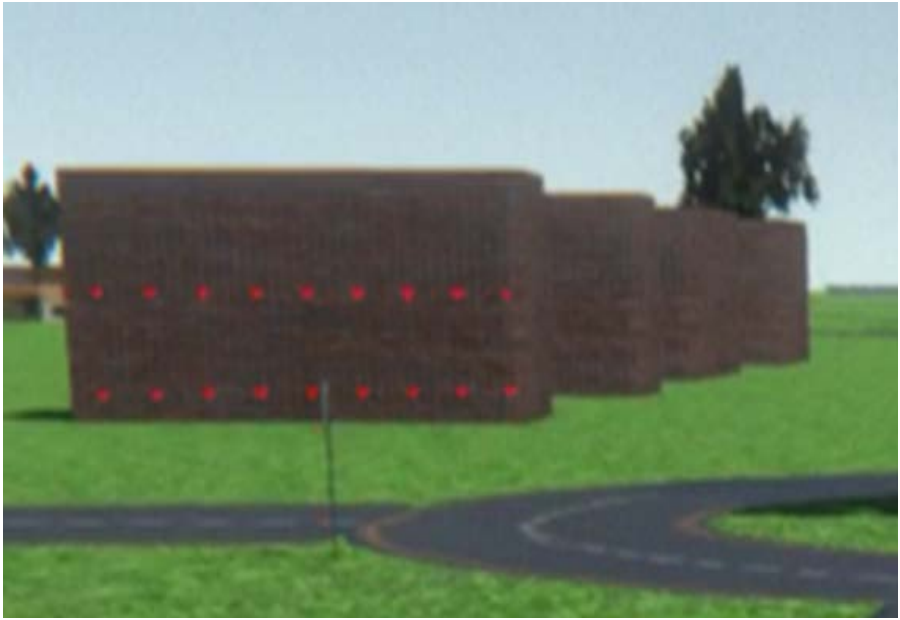

**Figure 17. Point Cloud Emulation in Unity based LG SVL Simulation**

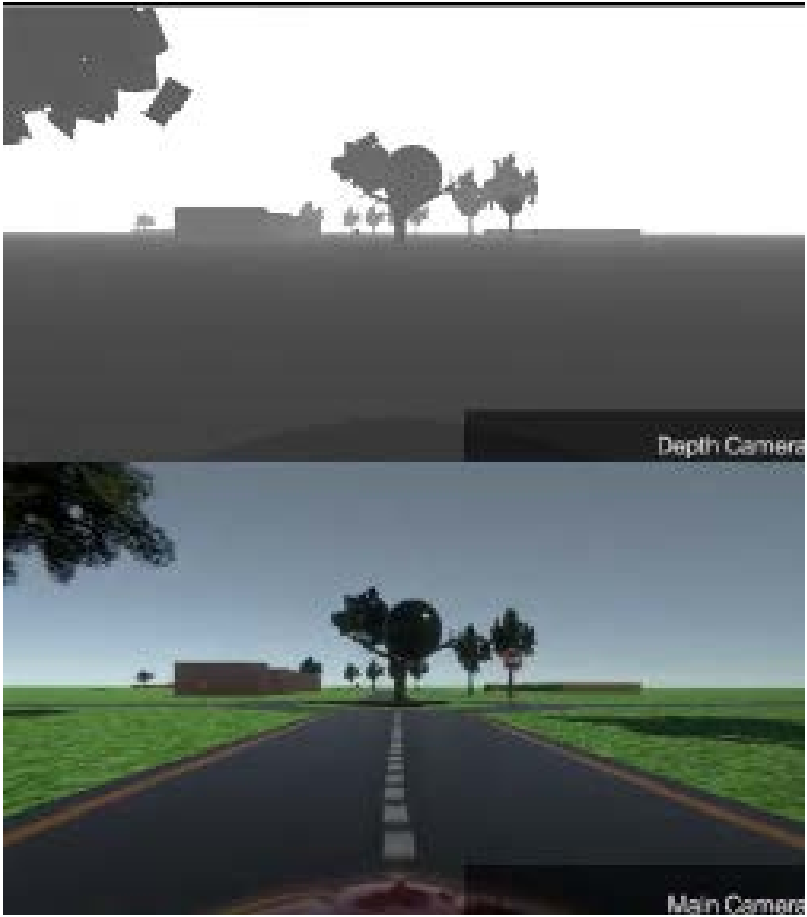

**Figure 18. RGB Camera and Depth Camera in Unity Based LG SVL Simulator**

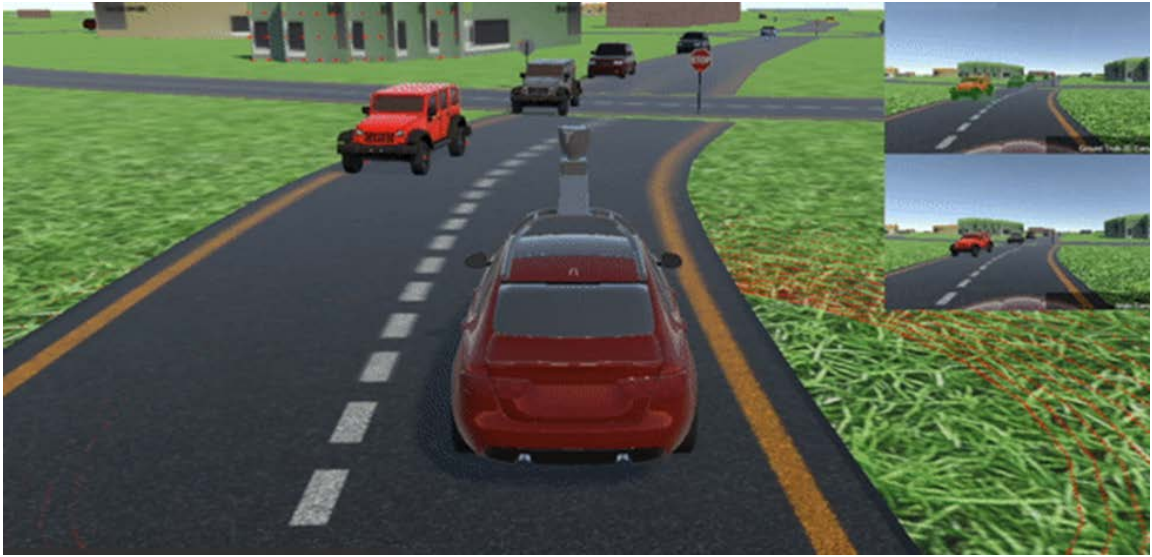

**Figure 19. Simulation of Traffic Using NPC Vehicles in Unity**

## 3.2. UNREAL ENGINE BASED SIMULATION

The freely available and open source CARLA simulator is used with the Unreal Engine platform simulation environment. CARLA has a powerful Application Programming Interface (API) that allows users to control all aspects related to the simulation including traffic generation, pedestrian behavior, weather conditions and sensors. Users can configure diverse sensor suites including lidars, multiple cameras, depth sensors and GPS for the AV. Within the simulation, ego vehicle's (AV) state of speed, heading angle, global location are easily seen using the virtual GPS sensor and states of the vehicle actuators. Those synthetic data can be accessed from other programs or software through the APIs provided by the CARLA simulator. Figure 20 shows the CARLA simulation of an AV in the Linden Residential Area environment. Figure 21 shows the corresponding camera depth map and Figure 22 shows a 2D lidar scan in CARLA.

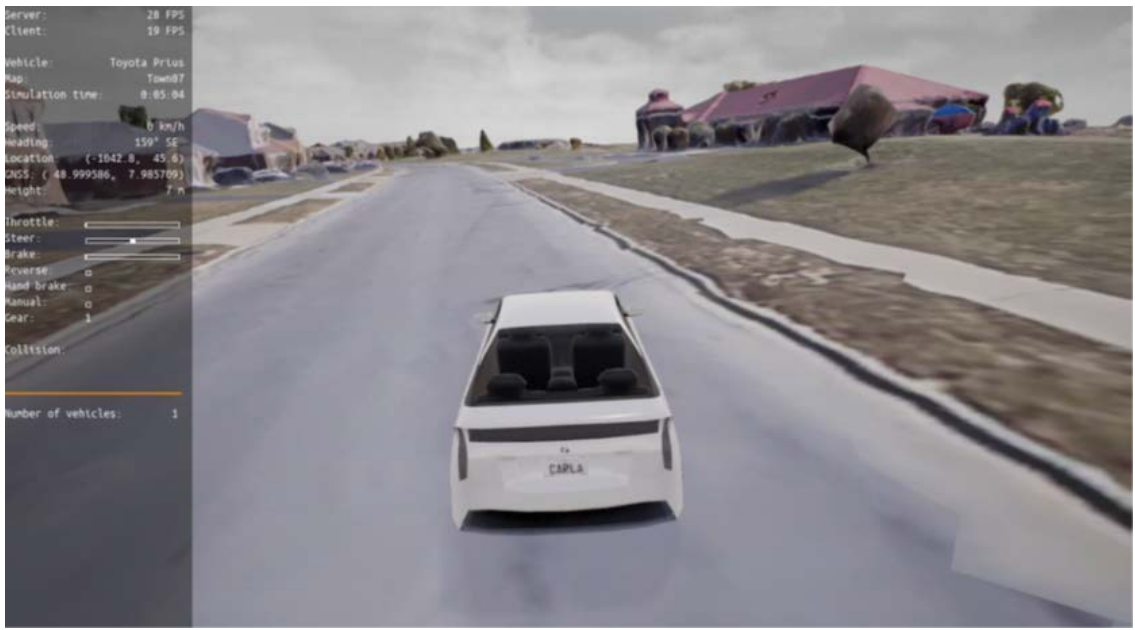

**Figure 20. CARLA AV Simulation in the Linden Residential Area**

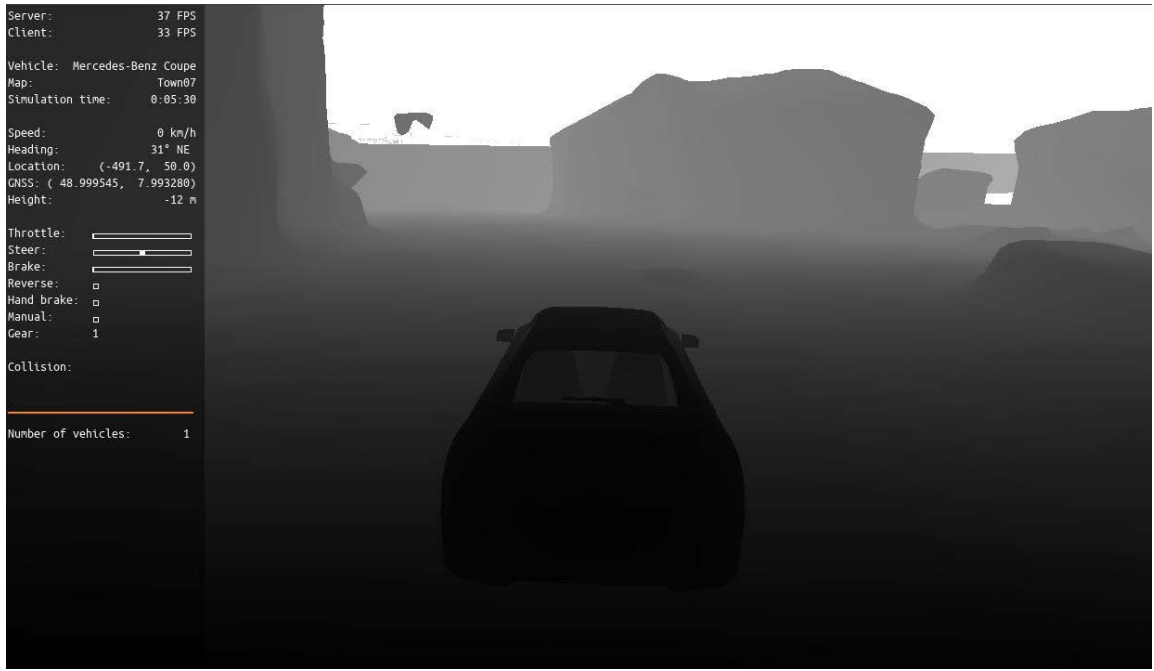

**Figure 21. Depth Camera Display in the CARLA Simulator**

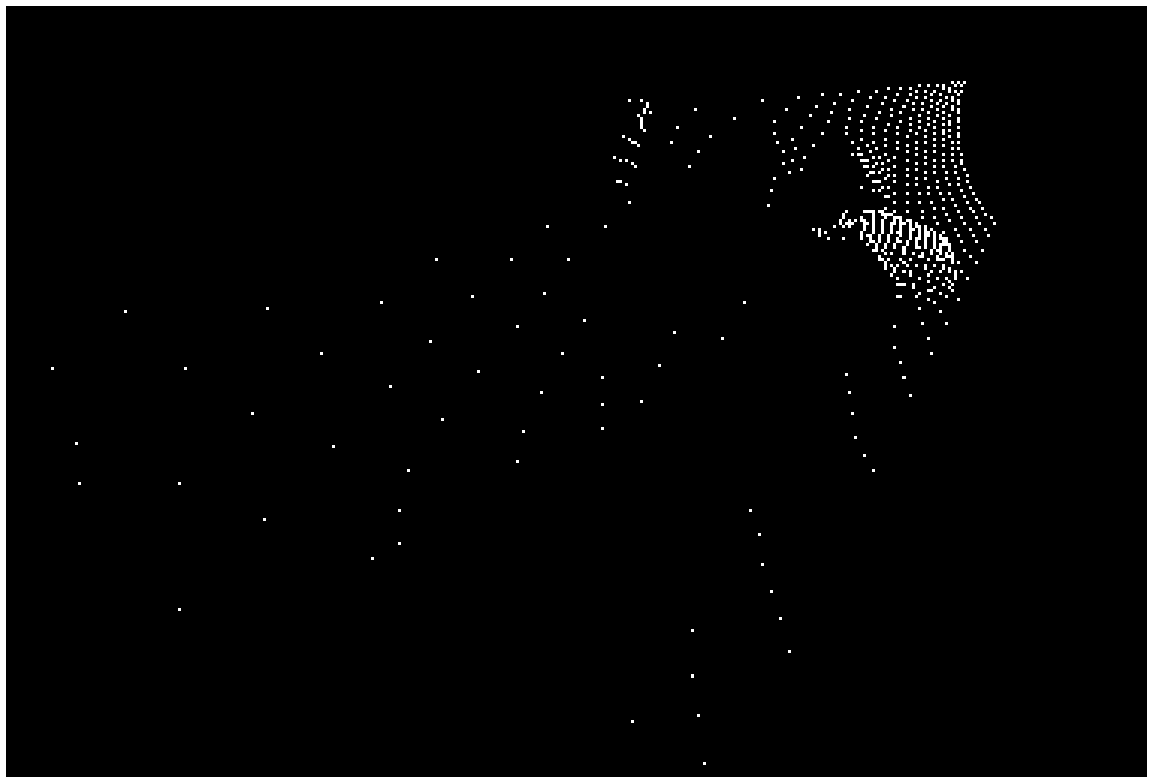

**Figure 22. 2D Lidar Scan in the CARLA Simulator**

## 3.3. CAPABILITIES OF THE SIMULATORS

Some of the important capabilities of the Unity and Unreal Engine platform based LG SVL and CARLA simulators that are significant in AV simulation testing in the Linden Residential Area are listed below.

- The virtual sensor suites in these simulators emulate the sensor suites deployed on current autonomous vehicle platforms, providing the capability of localization and perception in autonomous driving simulation.
- Multi-agent features in these simulators are used to generate different traffic scenarios of interaction with other NPC vehicles.
- The easy to use interfaces for these open source simulators make it possible to connect with external software and operating systems such as the Robot Operating System (ROS) so that readily available autonomous driving algorithms and functions in different software and operating systems can be tested and implemented.
- Additionally, it is easier to edit the simulation environment and generate different traffic scenarios in the open-source platforms.

Considering the many advantages of using these open source simulation tools, soft autonomous driving simualtions in the Linden Residential Area were prepared using them and is presented in the next chapter.

# Chapter 4. Autonomous Driving Simulation in the Linden Residential Area

In this chapter, the open source Autoware autonomous driving functions are used with the two open source simulators of the previous chapter to evaluate AV shuttle operation in the Linden Residential Area. Autoware [13] is an open source software for autonomous driving technology. It integrates a lot of autonomous driving functions based on ROS including functions for vehicle localization, object detection using different AV perception sensors and prediction and planning for autonomous driving. The open source Autoware approach is used here as AV companies do not share their autononmous driving algorithms. Most of the ROS routines that Autoware accesses are based on well established algorithms that have been used and tested extensively in the academic and research AV community. The motivation of the approach presented in this chapter is that the open source Autoware routines will be replaced by black box versions of similar algorithms by AV shuttle vendors to demonstrate and test their AV driving system before deployment on public roads. The Autoware and ROS routines used for autonomous driving simulation in this chapter are Normal Distributions Transform (NDT) mapping of the 3D point cloud map of the geo-fenced Linden Residential Area route, NDT matching for localizing the ego AV and a simple pure-pursuit algorithm for AV path following.

## 4.1. AUTONOMOUS DRIVING FUNCTIONS

### 4.1.1. Localization of the Ego Vehicle

Even though a high accuracy GPS unit can provide accurate global position information, its relatively high price and the presence of GPS-denied environments have prevented it from being a universal solution for the localization of autonomous vehicles. A cheaper and more widely used solution is based on comparing data from perception sensors like camera or lidar against the known locations of stationary landmarks. In mobile robotics applications which are mostly indoor, the Simultaneous Localization And Mapping (SLAM) method is used to determine the map, the walls, in an indoor environment and to localize the robot relative to that map by determining its correct position and pose with respect to the map. While the SLAM method can also be applied to autonomous driving, it is not fast enough for public road driving tasks. It also does not make sense to build the map again, each time the AV is running, because the map does not change drastically over time. For this reason, a map building followed by map matching method is preferred in AV applications. Navigation companies have already built high accuracy lidar point cloud based maps of most highways in the US with lane level positioning accuracy. Corresponding maps of urban areas are currently missing. However, this is expected to change soon as automotive OEMs and ride-sharing companies have started mapping urban roads also. It is also not very difficult to build an autonomous driving map in a geo-fenced area like the Scioto Mile Smart Circuit or the Linden Residential Area for university researchers and for the relatively small AV shuttle companies. Accordingly, the localization approach taken here also uses the same approach. The 3D lidar point cloud map of the Linden Residential Area was constructed to build a map that can be used for map matching based localization as our autonomous vehicle is driven through that route. The same approach is taken in the simulator environment by running a soft lidar to collect point cloud data and prepare a map which is then used with a map matching algorithm during the AV simulations for localization of the AV, i.e. its location and pose in the map.

Just like the freely available, open source and user contributed OpenStreetMap approach, it will be worthwhile for users to contribute to simulation environment maps and point cloud maps of urban roads and maybe highways later on, so that the research community can have access to these shared maps. The amount of data in these user contributed and updated simulation environments and AV driving maps is extremely large, making it hard to manage them but this is still a worthwhile endeavor. The Automated Driving Lab at the Ohio

State University has been working on this for many years by concentrating on possible urban AV routes in and around Columbus and in the OSU campus.

#### 4.1.1.1. 3D POINTCLOUD MAP BUILDING

The normal distributions transform (NDT) mapping algorithm applies the normal distributions transformation [14] on the lidar point cloud data, by first aligning the different frames of point cloud data collected by the lidar sensor and then transforming them into the same reference frame, followed by overlapping different frames of data together when they have similar probability density functions after the transformation. Normal distributions transform (NDT) operation for 3D lidar data is illustrated in Figure 23. A three dimensional (3D) point cloud is first gridded into multiple cells, then compute the normal distribution upon the point cloud within each of the grid. The 3D point cloud NDT algorithm works in similar manner by finding the transformation between the different frames of the 3D point cloud and aligning them in the same frame.

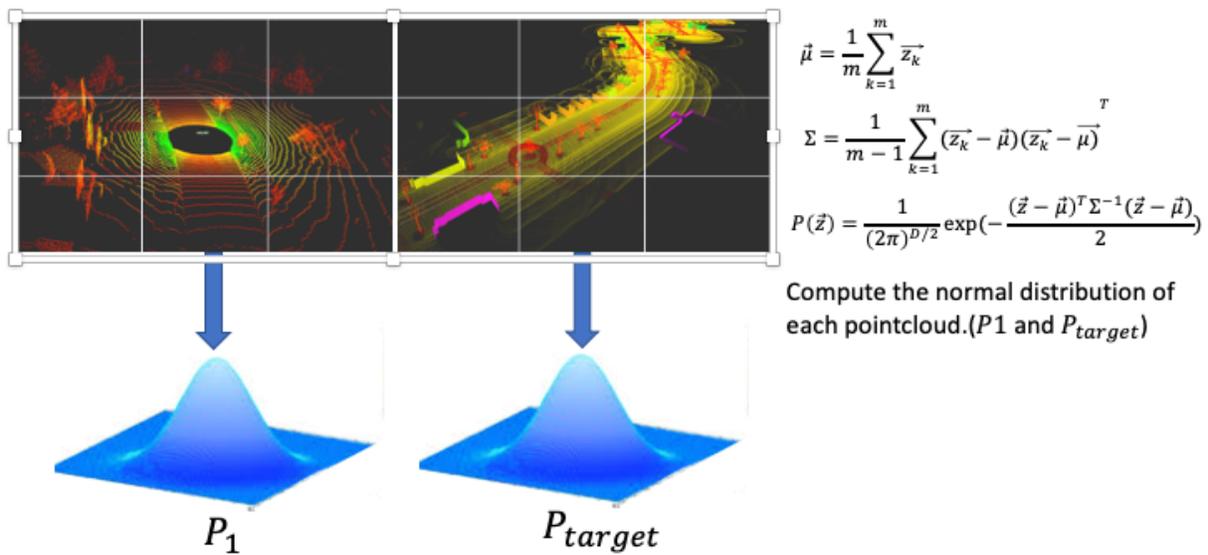

Compute the normal distribution of each pointcloud.($P1$ and $P_{target}$)

**Figure 23. Diagram of Normal Distribution Transformation Scan Matching**

#### 4.1.1.2. NDT MATCHING FOR LOCALIZATING THE EGO VEHICLE

Various algorithms of scan matching for AV localization in real-time have been developed for finding the transformation between two point clouds, such as Iterative Closest Point [15]. The first point cloud is the previously created map and the second point cloud is the lidar scan from the AV during its actual run. The approach used here is the normal distributions transform (NDT) method along with odometry extrapolation. For the NDT algorithm, the inputs are pre-built 3D point cloud map and the current point cloud scanned by the AV lidar Sensor. From the transformation information, the goal of localization in the reference frame of map is achieved. With odometry extrapolation, a better guess of initial transformation for matching is acquired so that the precision of matching between current scan and pre-built map is increased.

Here, the function $T(\vec{p}, \vec{x_k})$ is defined to be the space transformation function between the current point cloud data scanned by the lidar and the target 3D point cloud. The maximum likelihood function is used to find the pose $\vec{p}$ as

$$Likelihood: \Theta = \prod_{k=1}^{n} f(T(\vec{p}, \vec{x_k}))$$

The transform is found using optimization algorithms. Then, through the transformation between the lidar sensor of the AV, the space transformation is derived using the following equation:

$$\vec{p}_{ego} = g(\vec{p},\ T_{lidar})$$

to achieve localization by NDT matching.

### 4.1.2. Pure Pursuit for Path Following

Our aim in this report was to use a simple path following algorithm in the simulator for autonomous driving that can be replaced easily by user algorithms. The Pure Pursuit algorithm was chosen for this purpose as it is easy to implement, understand and tune and works well in low speed path following applications. Pure Pursuit [16] [7] is a geometric method of path following that follows a predefined path consisting of a series of waypoints. We first need to drive along the AV shuttle route to generate a series of path points with global information in the map and save them for later use as the path waypoints. In this report, the AV is driven along the Linden Residential Area simulation environment to record the waypoints. AVs operating on highways generate the waypoints automatically from a map. Once the waypoints are determined, the pure pursuit algorithm is applied for the autonomous vehicle to follow them on the autonomous shuttle route.

The bicycle model is used for modelling our vehicle in the pure pursuit algorithm. The geometric bicycle or single track model for a vehicle is illustrated in Figure 24, the radius of curvature during turning is simplified and represented by,

$$\tan(\delta) = \frac{L}{R}$$

where $\delta$ is the steering angle of the front wheel and L is the wheelbase with R being the turning radius when the vehicle is steering at a certain steering angle.

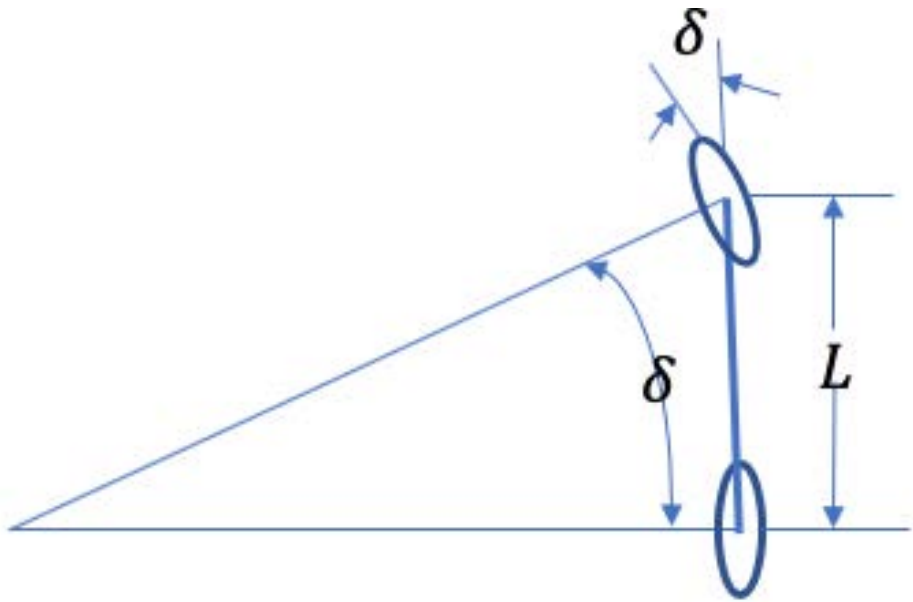

**Figure 24. Geometric Bicycle Model**

Pure pursuit aims to follow the next available waypoint as the goal point on the predefined path by controlling the steering angle. In the sketch of pure pursuit in Figure 25, the point with coordinate $(g_x, g_y)$ on the path is our next goal point and the vehicle always follows a trajectory that drives from current position to next way point as the goal points. The circular arc is the current path for the vehicle to reach the goal point when using bicycle model in pure pursuit algorithm.

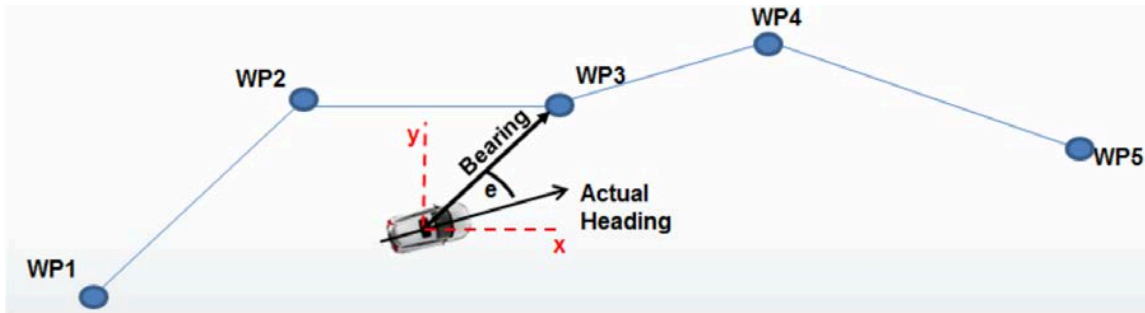

**Figure 25. Sketch of Pure Pursuit Algorithm Autonomous Driving Simulation**

The open-source traffic simulation tools we use have various scripts and APIs that can communicate with other systems and software. The communicating capability with ROS is used here [17] as illustrated in Figure 26. By transferring data according to Figure 26, multiple ROS routines for different AV driving functions can be implemented and used with the AV simulation generated sensor data. The freely available and open source Autoware autonomous driving library was used here due to the easier interface with ROS routines.

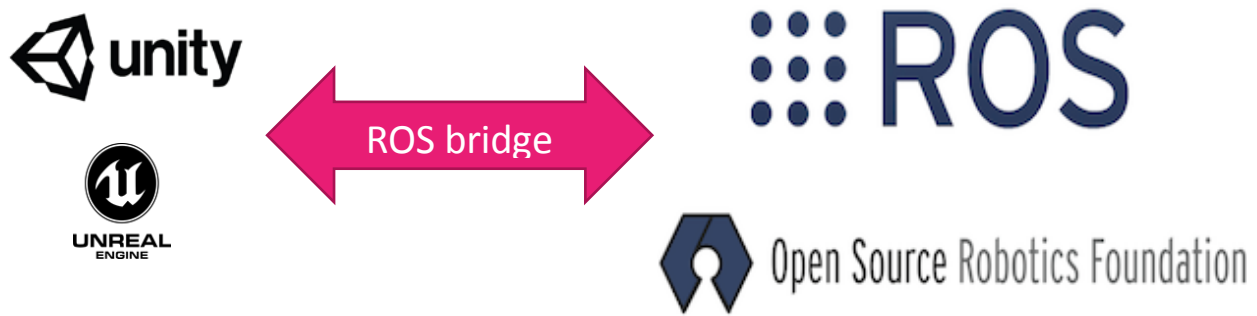

**Figure 26. Data Transfer with ROS During AV Simulation**

### 4.1.3. AV Driving Simulation on Unity Platform

This simulation concentrates on autonomous driving behavior along the Linden Residential Area autonomous shuttle route. Synthetic data is perceived by virtual sensors in the simulation environment that are sent to ROS for processing and the computed control commands are sent to the AV in the simulation. Sensor data during the simulations is visualized in ROS as shown in Figure 27 for lidar and Figure 28 for camera images.

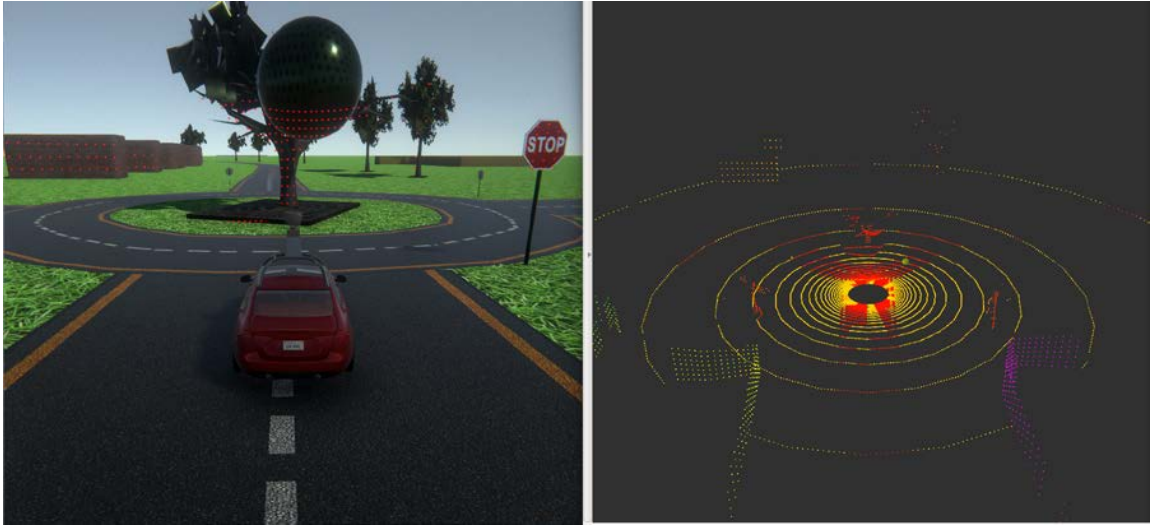

**Figure 27. Virtual Lidar Point Cloud in Unity Simulation (Left) and Visualization of Corresponding Point Cloud in ROS (Right)**

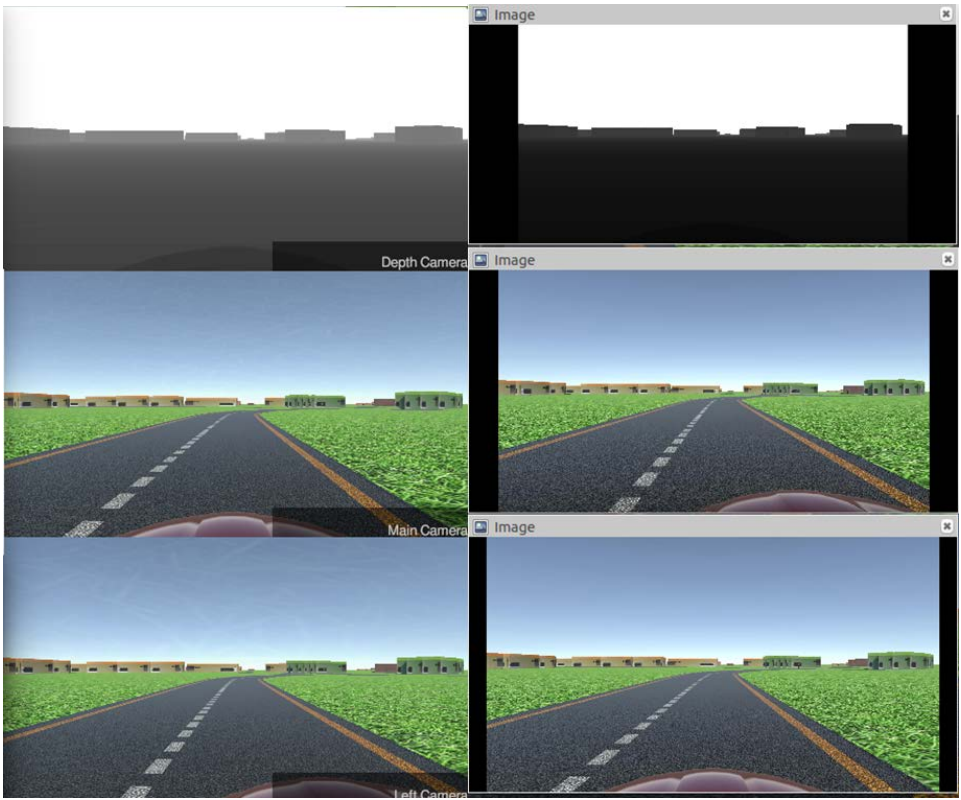

**Figure 28. Comparison of Camera Images in the Unity Simulation (Left) and Corresponding Image in ROS (Right)**

A 3D point cloud map of the Linden Residential Area AV shuttle route was built based on synthetic lidar data generated in the Unity platform LG SVL simulation using the NDT (Normal Distribution Transform) mapping technique that was introduced in the previous section. The point cloud map contains detailed environment information including road features, buildings and road signs that can be used for localization of the AV with respect to the map in the simulations.

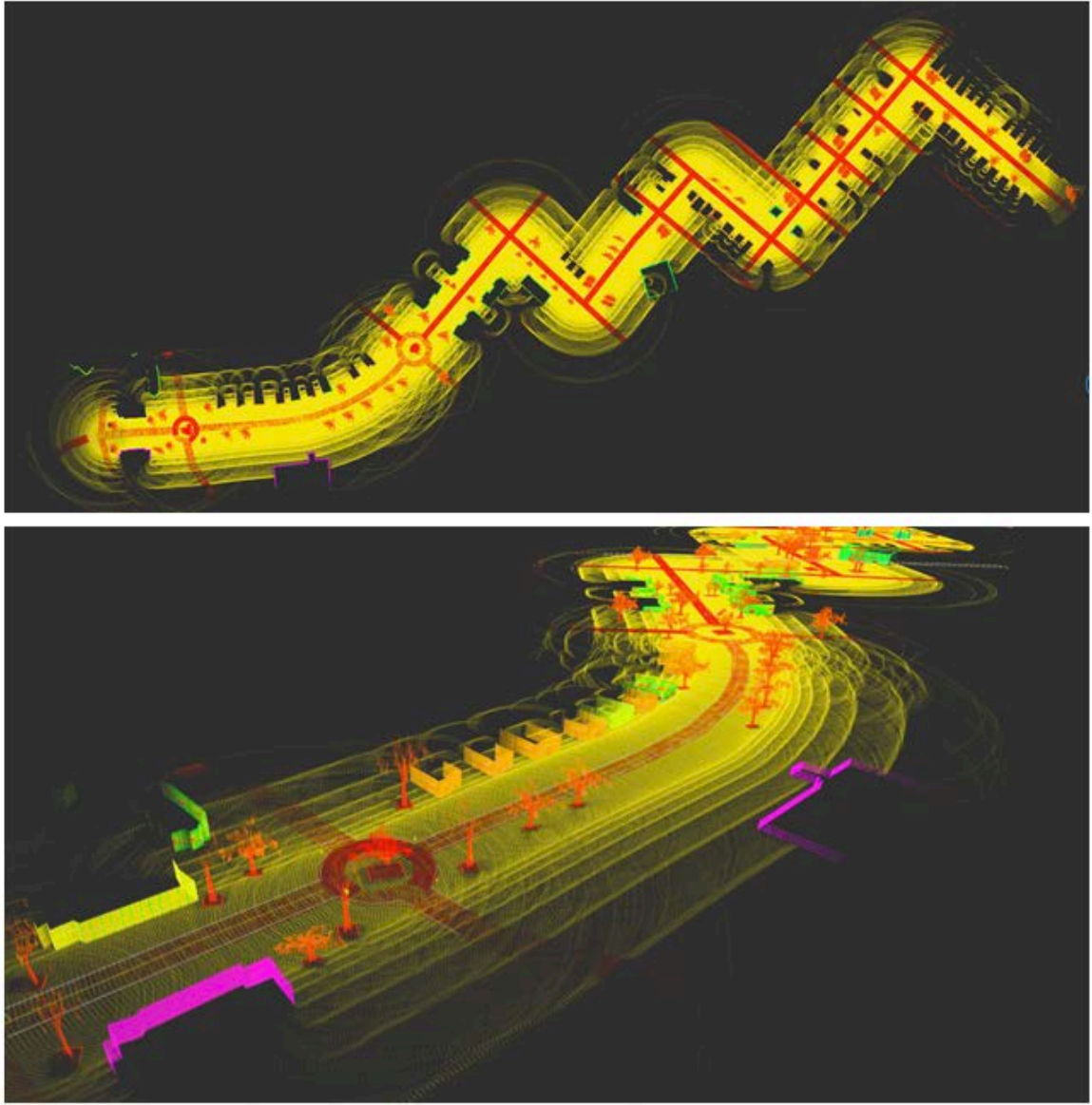

**Figure 29. Top View (Top) and Side View (Bottom) of the 3D Point Cloud Map of Part of the Linden Residential Area Route**

In the AV driving simulation, the NDT (Normal Distribution Transform) matching method is used for localizing the AV and for determining the AV's pose information. The way of using ndt-matching in Autoware is to enable the "ndt matching" tab from the "Computing" tab. In the AV driving simulations reported here, ndt-matching is enabled by using a launch file by clicking the "Localization" tab in "Quick Start" tab so that related functions are directly loaded within one file. It is more convenient to activate several function using a launch file in ROS as shown in Figure 30.

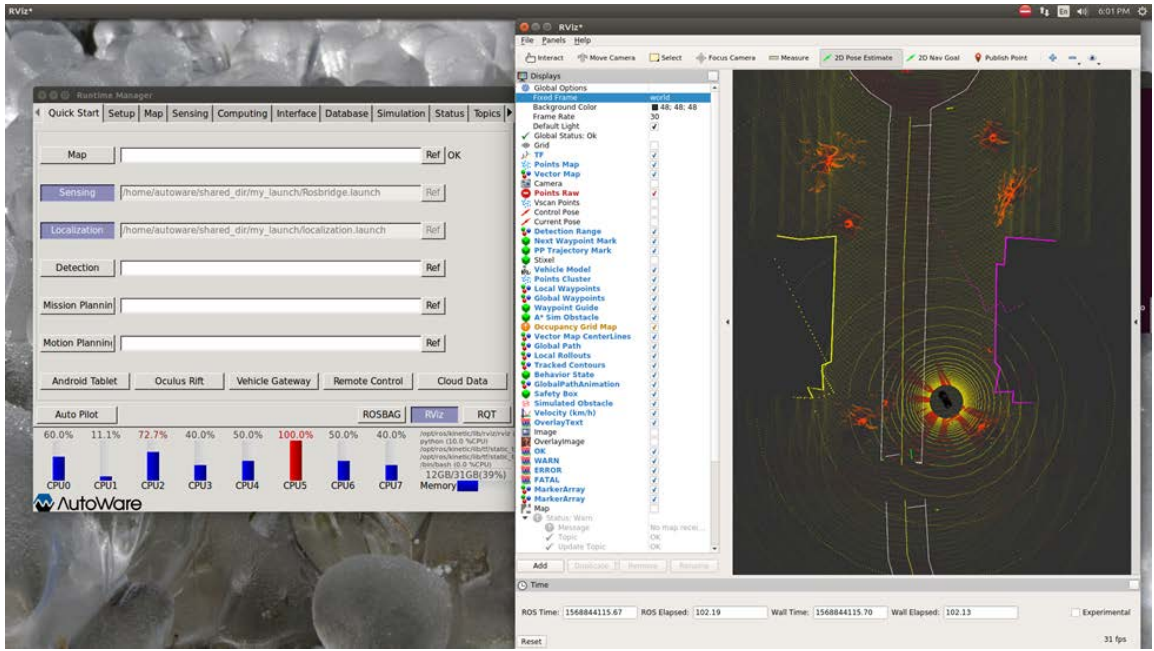

**Figure 30. Localization Activation and Initilization in AV Driving Simulation**

For initialization, the "2D Pose Estimation" function in ROS visualization package Rviz is used by pointing the arrow at the initial position for mapping of the AV. Every time, tracking for localization is lost in the simulation, one has to start from the beginning of the initial position of the AV. As shown in Figure 31, once the localization is successfully initialized, the current frame of lidar scan is aligned with the 3D point cloud map. Transformation information is computed when doing the alignment and used as localization information of the AV in the reference frame of the map.

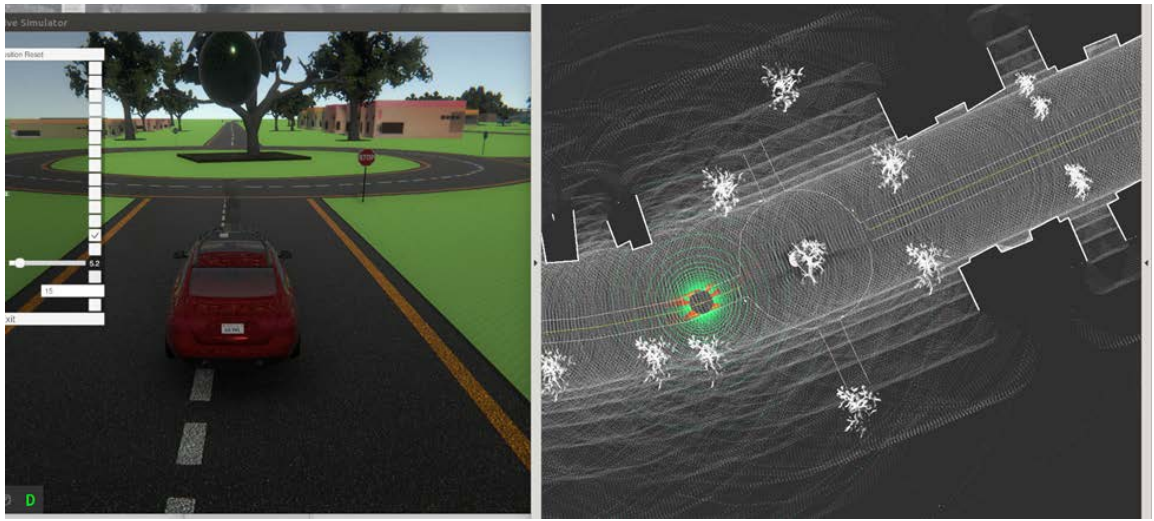

**Figure 31. NDT Matching Localization**

The waypoints along the autonomous shuttle route are generated and saved a priori by manually driving the simulated vehicle along the simulation route. The waypoints have position information based on the NDT matching and the relative frame of the coordinate is the map frame where the 3D point cloud map was built.

Then, a path is generated based on the saved waypoints in the map. In Figure 32, a series of waypoint is shown in the map.

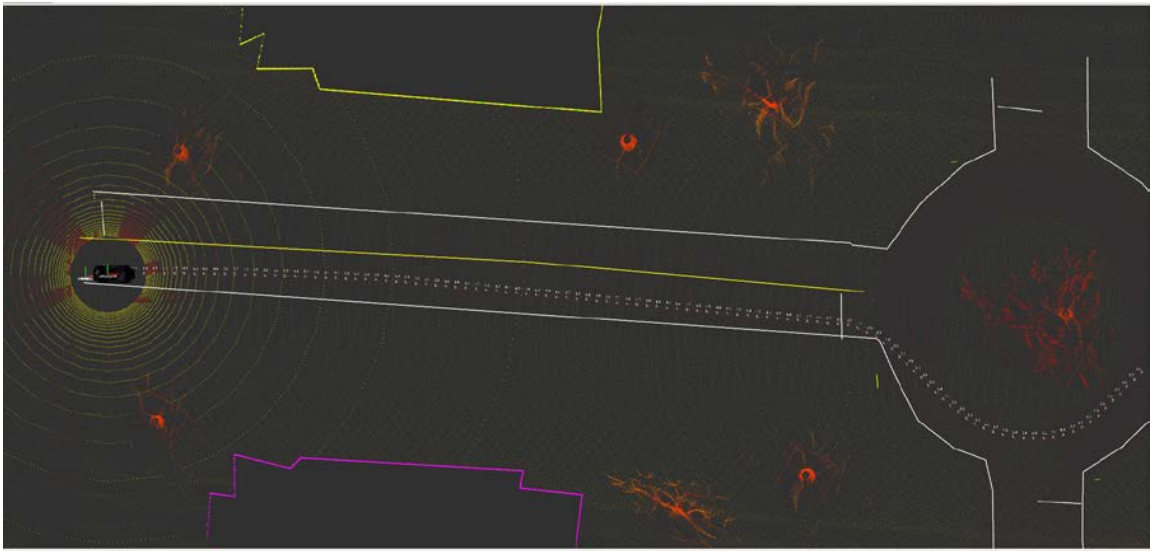

**Figure 32. Waypoints for Path Following in the 3D Point Cloud Map**

Then, with the waypoint loader in Autoware, all the predefined waypoints are loaded for path generation. Pure Pursuit is the function for autonomous path following algorithm for the AV driving simulations here. For implementation of the autonomous driving behavior, the following functions were enabled: lane_rule, lane_stop and lane_selection; Astar_avoid and velocity_set (A* algorithm) for route planning based on the loaded waypoints; Pure_pursuit and twist_filter for generating the steering command for the simulated AV. These functions are displayed in Figure 33.

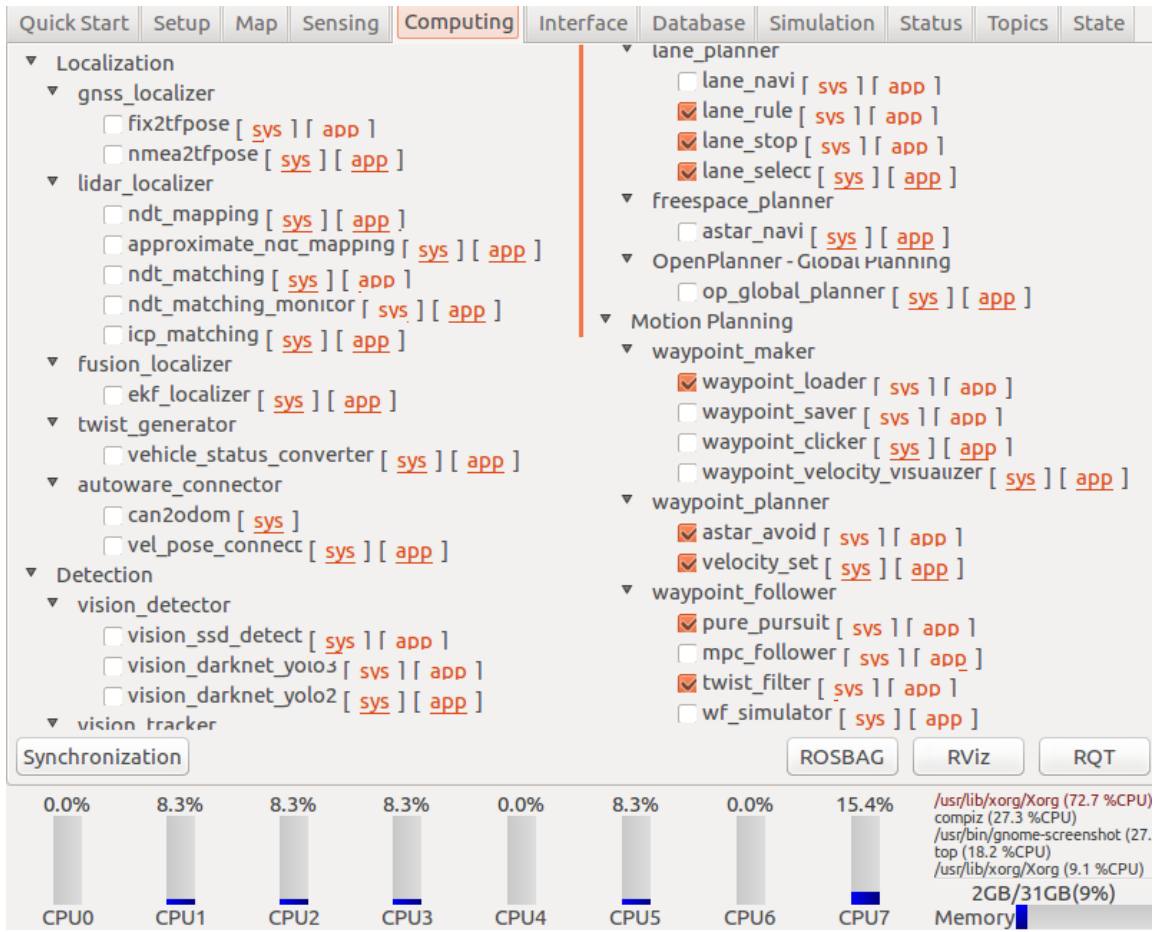

**Figure 33. Functions for Pure Pursuit in Autoware**

As is shown in Figure 34 and Figure 35, the bigger green point on the path is the goal point of current state and it is updated and generated with the change of the state of the AV during the simulation. The detailed demonstration video corresponding to this autonomous driving simulation along the Linden Residential Area AV shuttle route can be viewed at: https://youtu.be/2U_MmXpVnEw. An advantage of the AV driving simulation approach presented here is that the AV computations can easily be transferred to an NVIDIA system like the NVIDIA Drive PX2 and which can be used as hardware in HiL simulation and as the main AV control unit in an actual AV.

Linden Residential Area CEAV Simulation Evaluation – Final Report | Smart Columbus Program | 31

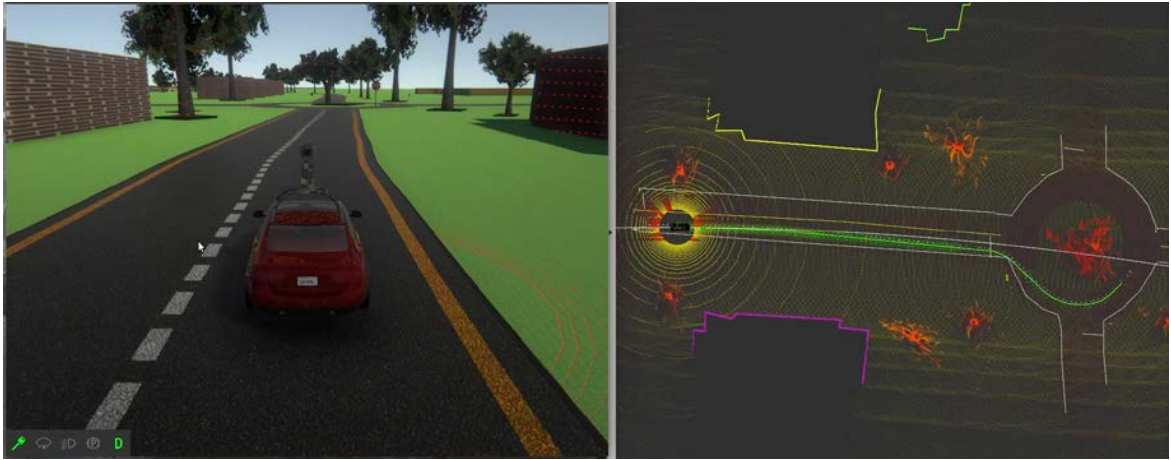

**Figure 34. Pure Pursuit AV Path Following along the Linden Residential Area Route**

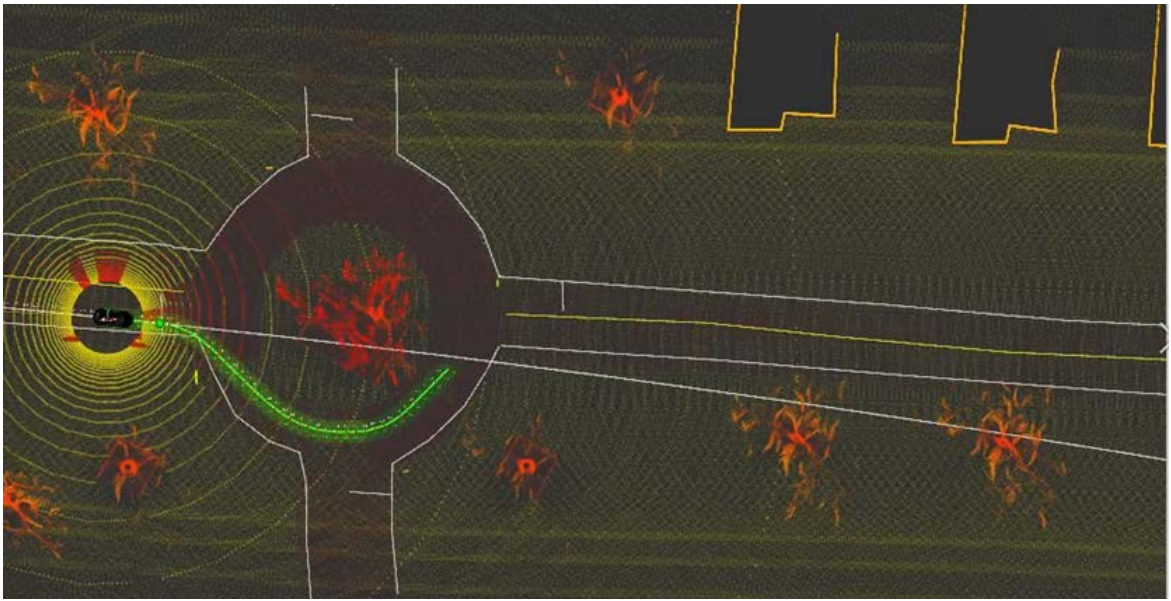

**Figure 35. Pure Pursuit AV Driving Simulation Visualization in ROS**

### 4.1.4. AV Driving Simulation on Unreal Platform

Similar AV driving simulation was also carried out in the Unreal Engine platform using the CARLA simulator. Autoware routines including Path Planner and Vector Map Navigation were used in the Unreal platform CARLA AV driving simulations as shown in Figure 36.

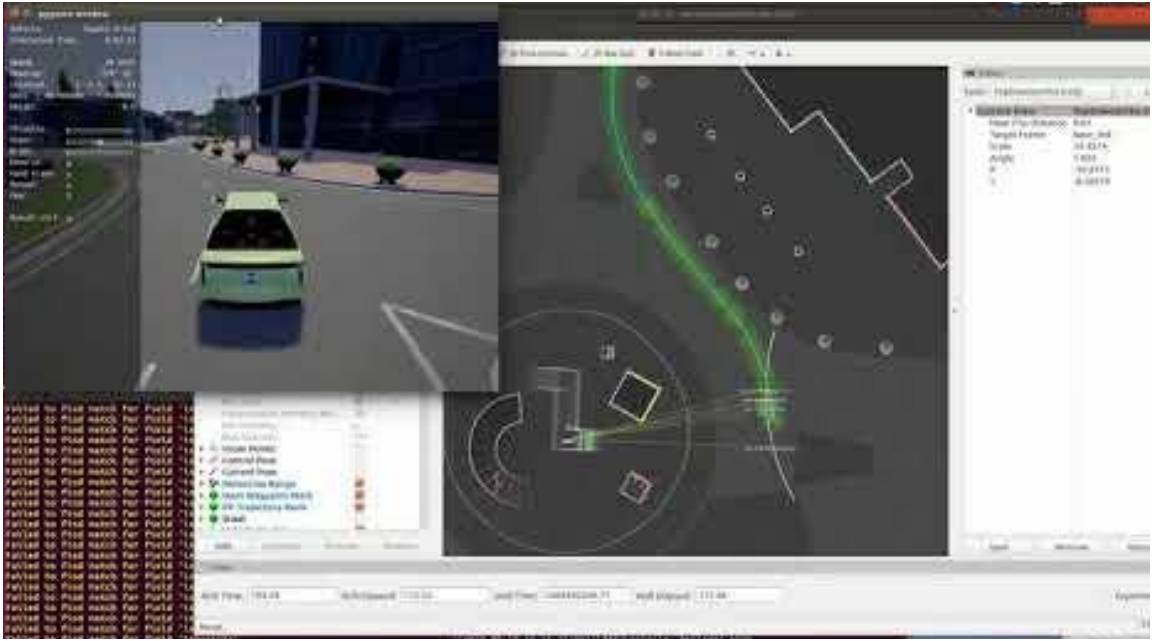

**Figure 36. AV Driving Simulation in Unreal Engine Platform CARLA Simulator**

## 4.2. OTHER WORK AND INTEGRATION INTO TEACHING

Most of the work reported here was based on the use of freely available code and programs. While this approach works well for the AV driving algorithms, the AV simulation environment construction process can benefit from the use of commercially available tools with automatic road generation and better rendering capabilities. One such tool for the environment generation is the program the Roadrunner [18] from VectorZero which makes it easier to create a realistic simulation environment from an OSM map. With Roadrunner, one can directly import Geographical Information System (GIS), lidar point cloud data and open source map information into the simulation environment. With better mesh generation and rendering, the simulation environment will be more realistic. An example of Roadrunner capabilities taken from their web site is shown in Figure 37.

Apart from the work and results present in this chapter, some other work is in progress and is explained briefly as an example of how the simulation environment and simulated AV driving method of this report can be used. These examples are current and planned work on how the simulation environment and AV driving simulation can be used in graduate student thesis research and in a course on Autonomy in Vehicles.

The first one of these examples of other work is the use of the simulator and AV driving simulation for developing a reinforcement learning based end-to-end AV driving system using the method of behavior cloning with affordance [19]. This ongoing work that supports graduate student thesis research focuses on end-to-end learning for automated driving in different traffic scenarios. A deep neural network is trained with data acquired by driving under different traffic scenarios in the simulation environment created. This behavior cloning approach to AV driving is illustrated in Figure 38.

Linden Residential Area CEAV Simulation Evaluation – Final Report  |  Smart Columbus Program  |  33

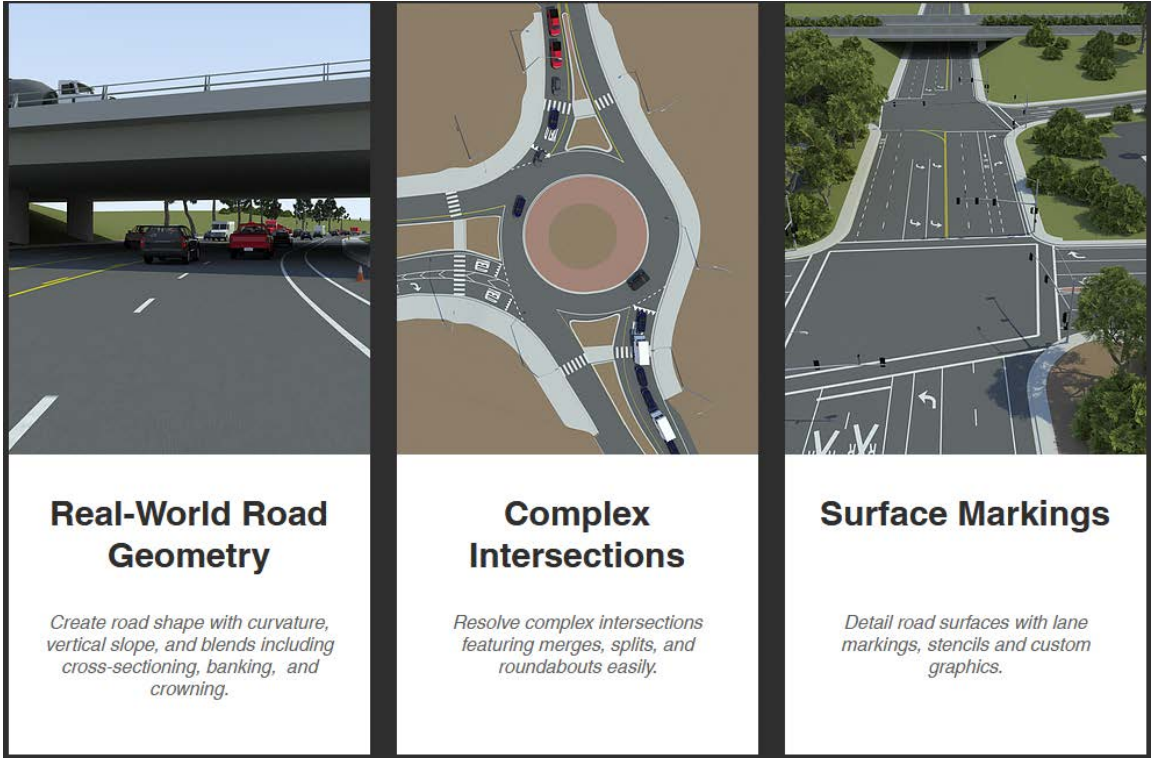

**Figure 37. RoadRunner Capabilities**

*Source: https://www.vectorzero.io/roadrunner*

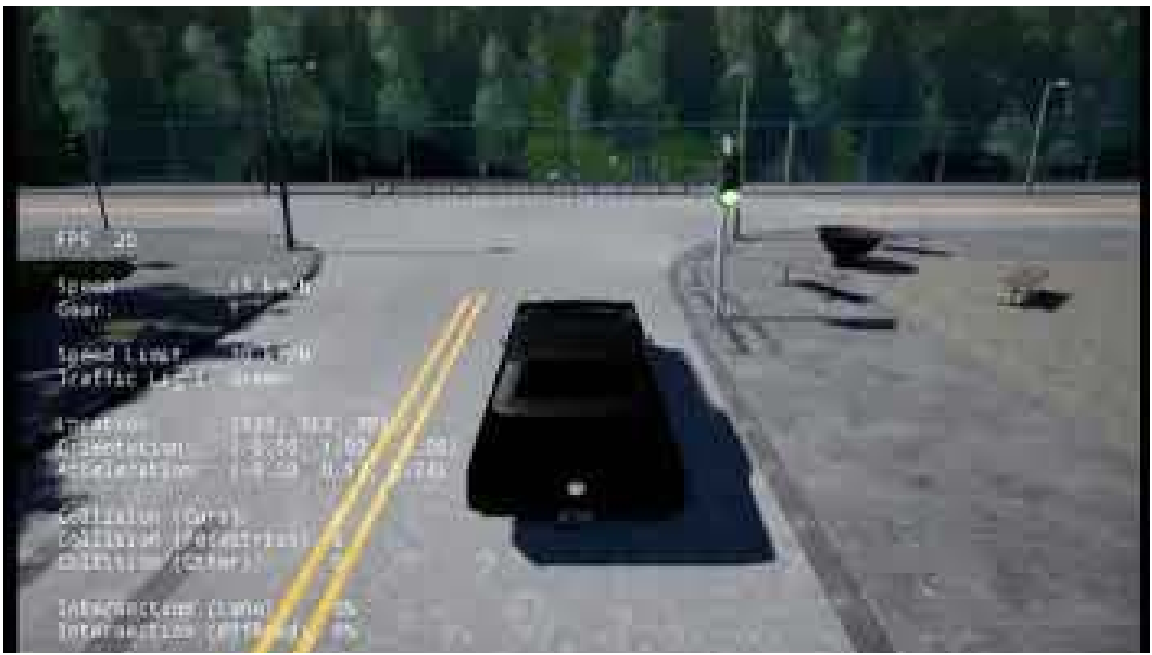

**Figure 38. The Interface for Behavior Cloning**

**34** | Smart Columbus Program | Linden Residential Area CEAV Simulation Evaluation – Final Report

Collision Avoidance based on the occupancy grid map and the A* path planning algorithm is also being investigated for safe AV driving using the simulation environment and AV driving simulation of this report in support of graduate student thesis research. As shown in Figure 39, object clustering based on lidar detection is carried out and the objects' position information is given by the Euclidean Clustering Algorithm in the simulation environment of this report.

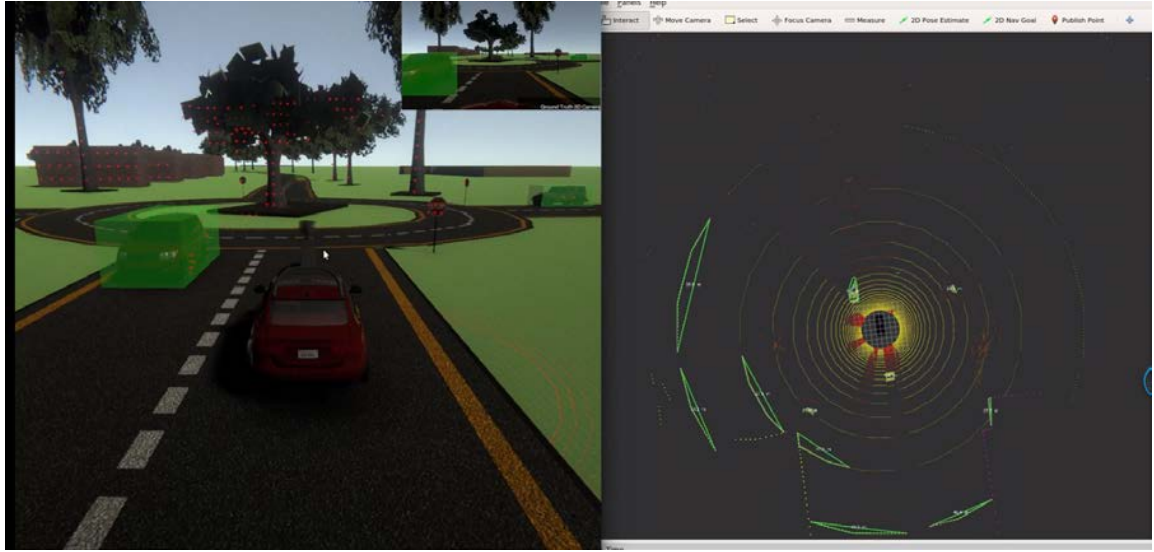

**Figure 39. Clustering Based on Lidar Scan based Detection**

Based on the clustered objects' position information, an occupancy grid map is generated that classifies the area around the ego vehicle into two parts: the walkable area and the non-walkable area based on which collision avoidance algorithms can be used. In Figure 40, the generated occupancy grid map is shown. The white block is the area under detection. The black region means that objects or obstacles are detected where the region is classified as a non-walkable region and the white region in the detection area is the walkable region since no obstacle is detected.

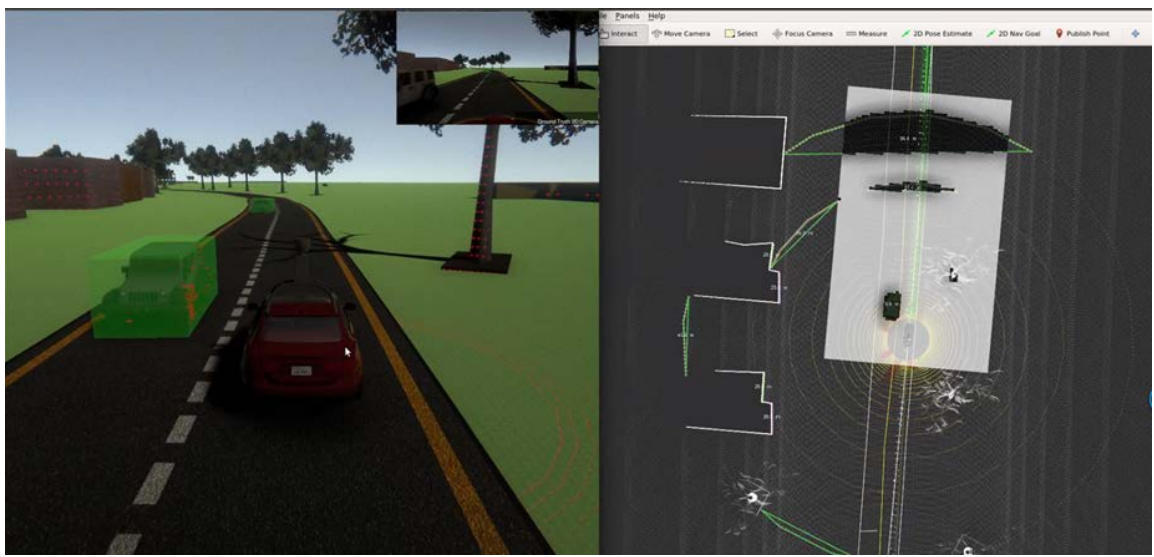

**Figure 40. The Occupancy Grid Map Around the Ego-vehicle**

The Linden Residential Area AV simulation environment and AV driving simulation approach of this report will be incorporated into the teaching material of the Ohio State University Electrical and Computer Engineering Department course ECE 5553 Autonomy in Vehicles during its next offering on Spring 2020. The initial and final projects of the course will be re-designed and will be based on a low speed AV shuttle operating along the Linden Residential Area route. This course is taken by both graduate and higher level undergraduate students and has an attendance of about 65 students most of whom are interested in a job in the AV industry. Most of the students were from the Electrical and Computer Engineering Department and the Mechanical and Aerospace Engineering Department of the Ohio State University. Up to 15 practicing engineers working in automotive OEMs have also taken part in the distance education section of this course in the past.

# Chapter 5. NVIDIA and CarMaker HiL Simulation on the Linden Residential Area Route

This chapter is on the use of the NVIDIA Drive PX2 autonomous driving computing hardware in HiL simulation of AV driving in the Linden Residential Area simulation environment. The NVIDIA Drive PX2 was chosen as the computing platform here as two NVIDIA Drive PX2 units that were donated by NVIDIA to the Automated Driving Lab of the Ohio State University were available for this work. Other computing platforms with the ability to run deep neural networks for perception and decision making algorithms such as the ones used in the actual AV shuttles in operation can also be used in HiL simulation evaluation to repeat the work presented in this chapter.

The commercially available CarMaker program was used as the simulator in the HiL simulations due to the ease of creating raw camera and lidar images that can be sent to the NVIDIA Drive PX2. The ease of interfacing CarMaker and Vissim for traffic simulation also made it easier to add realistic traffic into the simulations with this approach. It is possible to replicate this approach using the Unity LG SVL and Unreal CARLA simulators that were discussed in previous chapters. The overall system within the co-simulation environment here has three main components which are the 3D map, the traffic information, and the decision making of the AV. The traffic simulator Vissim only works on a Windows platform while development for the NVIDIA Drive PX2 is performed on an Ubuntu Linux platform. Thus, the co-simulation set up runs Vissim, CarMaker and the NVIDIA Driveworks perception and decision making algorithms on different machines with different operating systems. The data resulting from these processes were converged on an external storage device. Meanwhile, the flow of data from the CarMaker simulation and the NVIDIA DRIVE PX2 perception and decision making is shared only between those two systems. This architecture of information exchange during the simulations is shown in Figure 41.

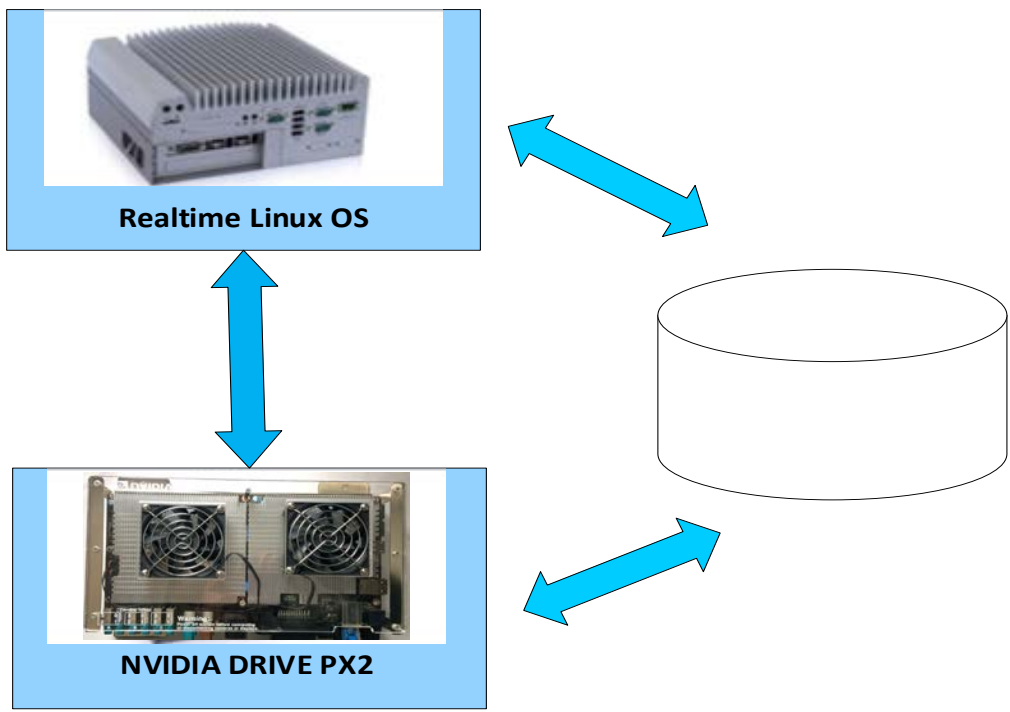

**Figure 41. System Architecture of Information Flow between NVIDIA Drive PX2 and CarMaker**

The data processing runs as a back-end process during the CarMaker simulation. A library for each sensor was written in C++ and incorporated into the CarMaker user code file User.cpp. The libraries act as an interface for spoofing data packets in order to transmit data in the correct format for NVIDIA Driveworks Dynamic Neural Networks (DNN). The advantages of this approach go beyond having the correct format. A faster interface than the Matlab/Simulink interface is achieved. Since the sensors are modeled based on physics principles, their simulated operation is more realistic, and it is relatively easier to introduce sensor.

The format that CarMaker uses to define maps is native to it. That is, in order to edit or create the map, one must directly modify the text file contents or use the provided Scenario Editor GUI. This poses a challenge since many changes through the GUI editor tend to create crashes. The strategy used was to import an OpenStreetMap section to an OpenDrive file and then to import it into CarMaker to obtain a blueprint. On this blueprint, the roads and intersections were fixed for the Linden Residential Area route. More often than not, this means redefining the entire network. The Linden Residential Area map was created in CarMaker using this approach and after defining the roads and intersections; vegetation and buildings were added manually. It is worth noting that the map creation for town areas is a difficult task. Furthermore, research has been done on collecting high-definition maps for AV simulation, but a couple of challenges arise. The first, is that there is no single unanimous format file for such environments. Different format files specify different levels of details on the maps. As the maps we require are for AV simulation, we look for maps that are detailed enough for our sensors to be able to make a decision but not too detailed as to slow down the simulation. The second issue is that while high definition maps for AV driving do exist, these maps have traditionally concentrated on highways. This does not only mean that the study of urban areas is limited but also that the available maps may not objects such as buildings which are important for emulating map matching based localization.

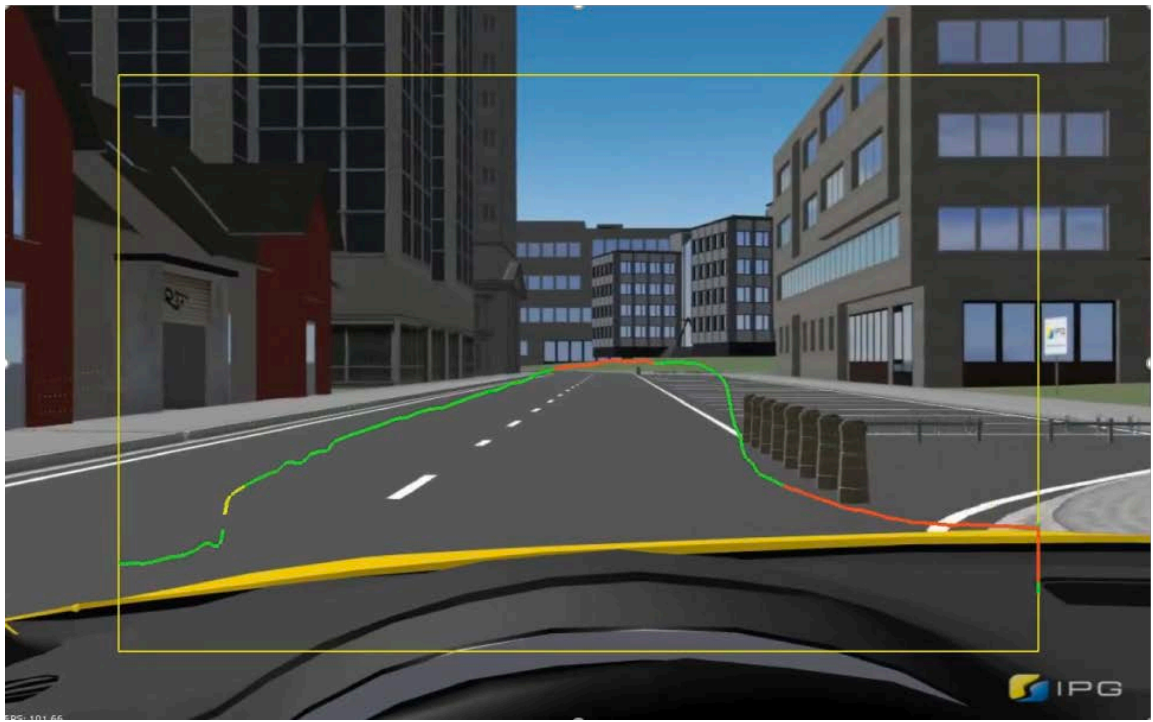

**Figure 42. Nvidia Driveworks Free Space Detection Algorithm**

The NVIDIA Drive PX2 platform is an architecture intended to be used as an in-vehicle PC and aims to replace any other interfaces between the Inputs/Outputs (I/O) of the vehicle. The Drive PX2 comes with the Driveworks software for data processing and decision making. The Driveworks software available for the Drive PX2 is made of three types of modules which are the Sensor Abstraction Layer, Algorithm Modules and Perception DNNs

38 | Smart Columbus Program | Linden Residential Area CEAV Simulation Evaluation – Final Report

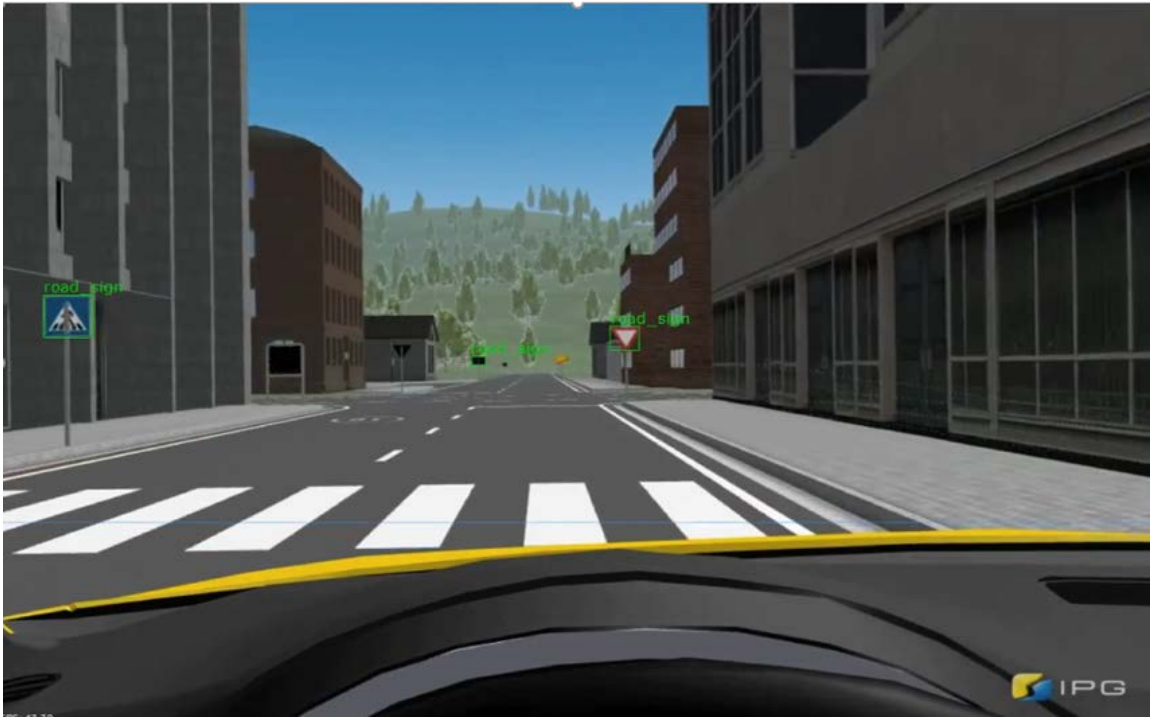

**Figure 43. Nvidia Driveworks Object Detection and Classification Algorithm**

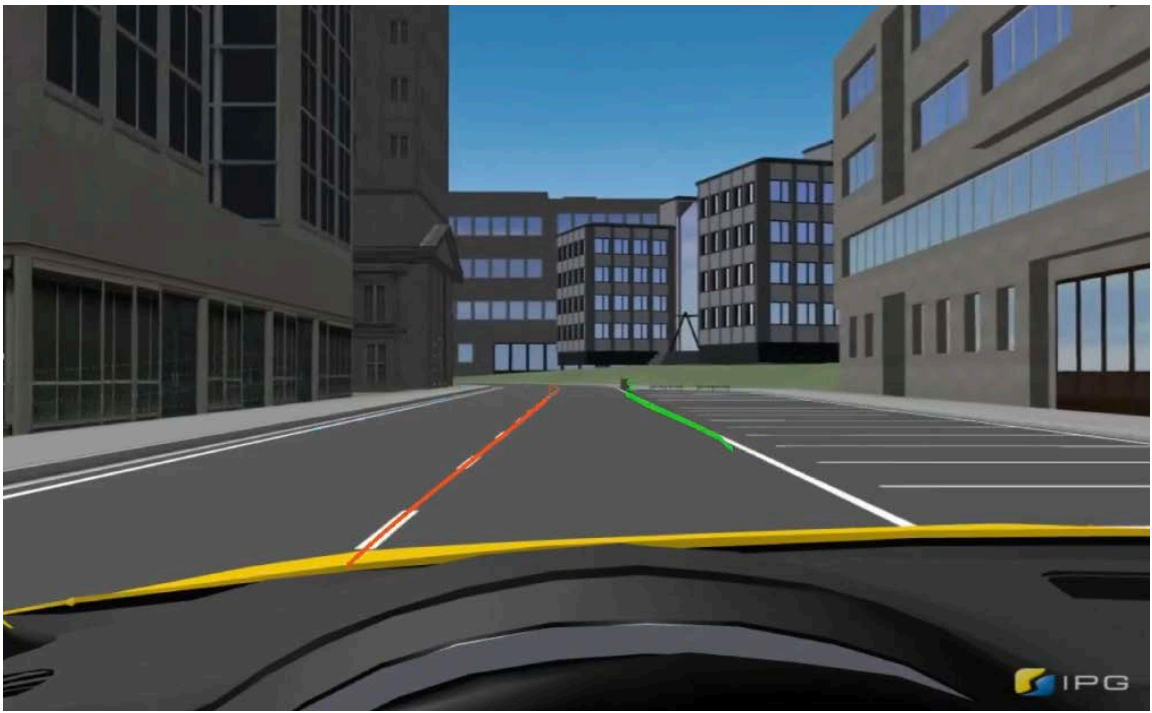

**Figure 44. Nvidia Driveworks Lane Detection Algorithm**

In the implementation reported here, the camera data generated from the CarMaker simulation is sent to the NVIDIA Drive PX2 which runs the corresponding Driveworks algorithms at each simulation step. Figure 42 shows a screen shot of the Driveworks freespace algorithm working during the simulation. Figure 43 shows a screen shot of Driveworks object detection and classification algorithm during the simulation. Figure 44 shows a screenshot of the Driveworks lane detection algorithm in operation during the simulation.

# Chapter 6. Safety Assessment Using AV Simulation Environment and Tools

A significant motive behind developing realistic AV simulation environments and AV driving simulation is to evaluate the safety and reliability of an AV before public road deployment. AV driving simulation allows developers and public safety agencies to do extensive in the lab and thus safe testing before moving on to controlled area testing and preliminary public road evaluation. AV driving simulation also allows the easy testing of unsafe collision possibility events and weather conditions which are unsafe or difficult to reproduce in a repeatable fashion during road testing. Simulation environments and simulators for AV driving like the ones discussed in the previous chapters of this report need to be complemented with AV safe operation testing cases which need to be updated and augmented over time. AV simulators like the ones presented in this report also make it easier to evaluate AV safety and performance during operation in realistic traffic including rare, unexpected events. An example of the use of AV simulators for testing how an AV handles unexpected situations is presented in the CARLA Challenge [20] The Linden Residential Area simulation environment was converted into a stand-alone LG SVL simulation by changing it into executable code. This procedure has taken on the order of one day of computer use in the case of this report but the result achieved is worth it. The AV research user community can now access this environment in the freely available and open source LG SVL simulator and try their own AV driving algorithms and test against challenging events.

In the AV simulator for the Linden Residential Area, several different challenges are defined as handling an unsignalized intersection with random other traffic as shown in Figure 45, handling a traffic circle with random other traffic as shown in Figure 46, handling the sudden appearance of a pedestrian as shown in Figure 47. Other test cases under consideration and development are the sudden detection of a stopped vehicle as shown in Figure 48 and interaction with a pedestrian as shown in Figure 49.

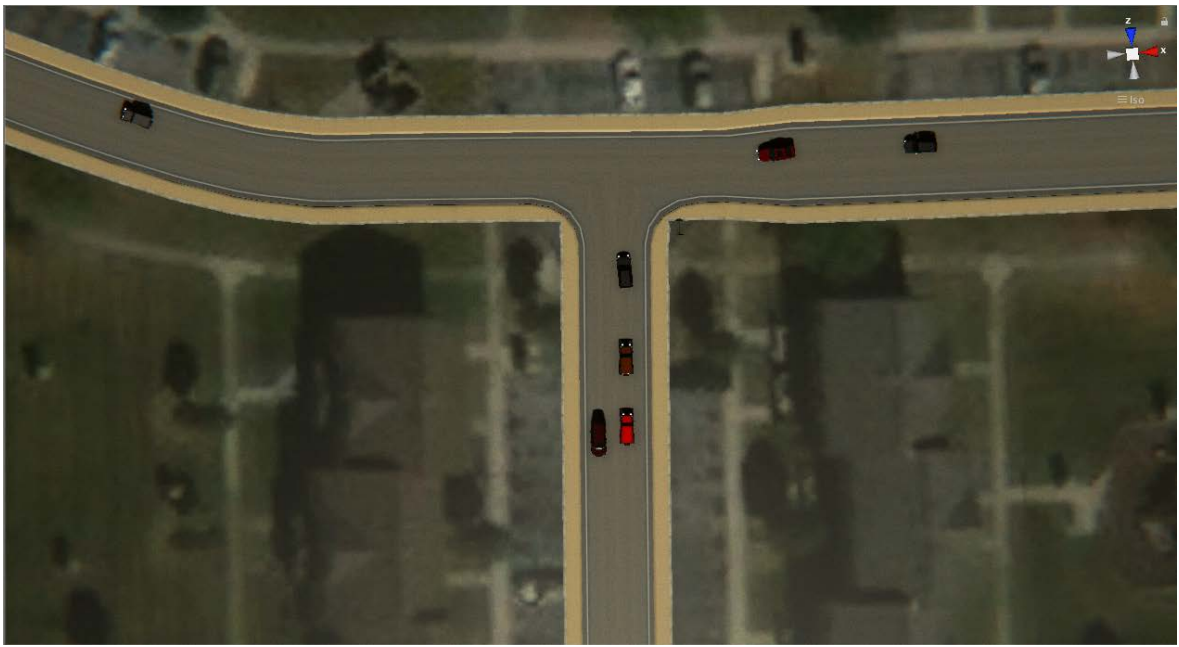

Figure 45. AV in Unsignalized Intersection with Random Traffic

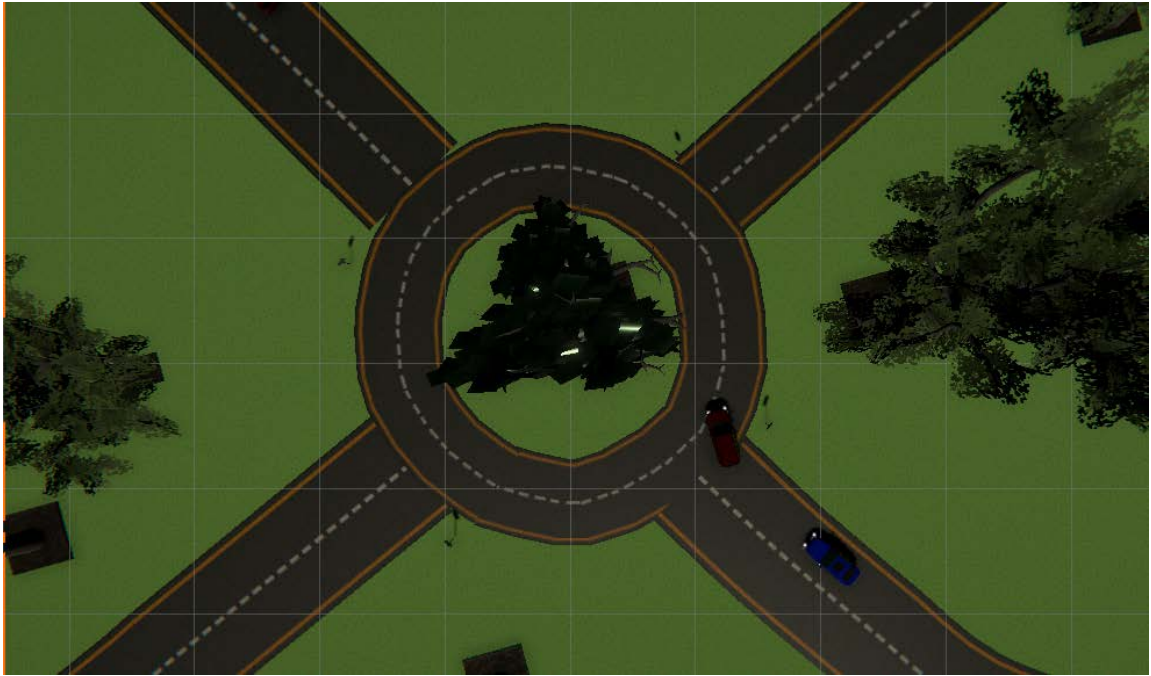

**Figure 46. AV in Unsignalized Traffic Circle with Random Traffic**

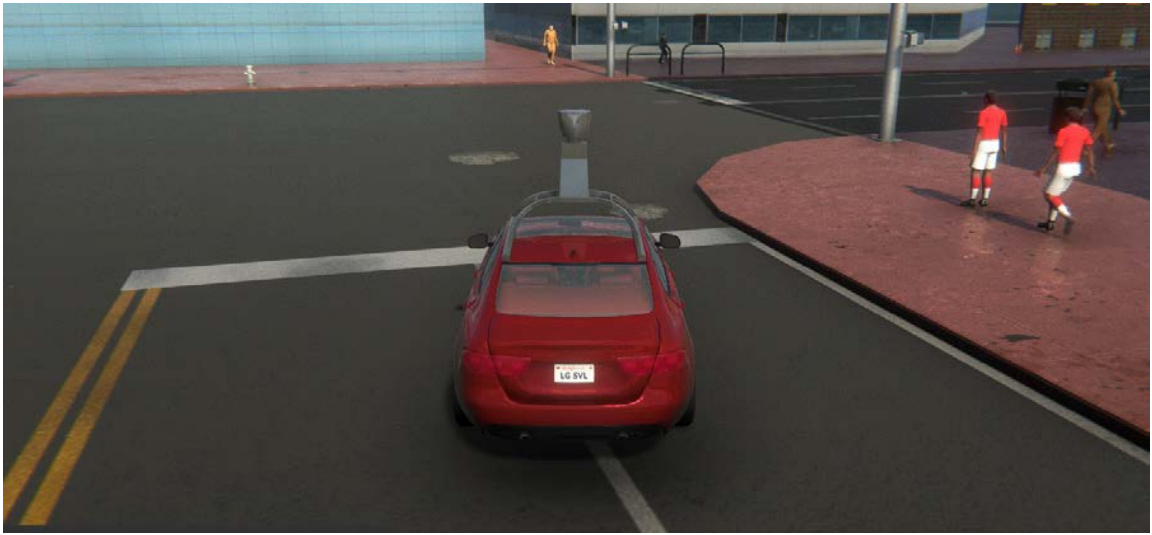

**Figure 47. AV in Pedestrian Crossing**

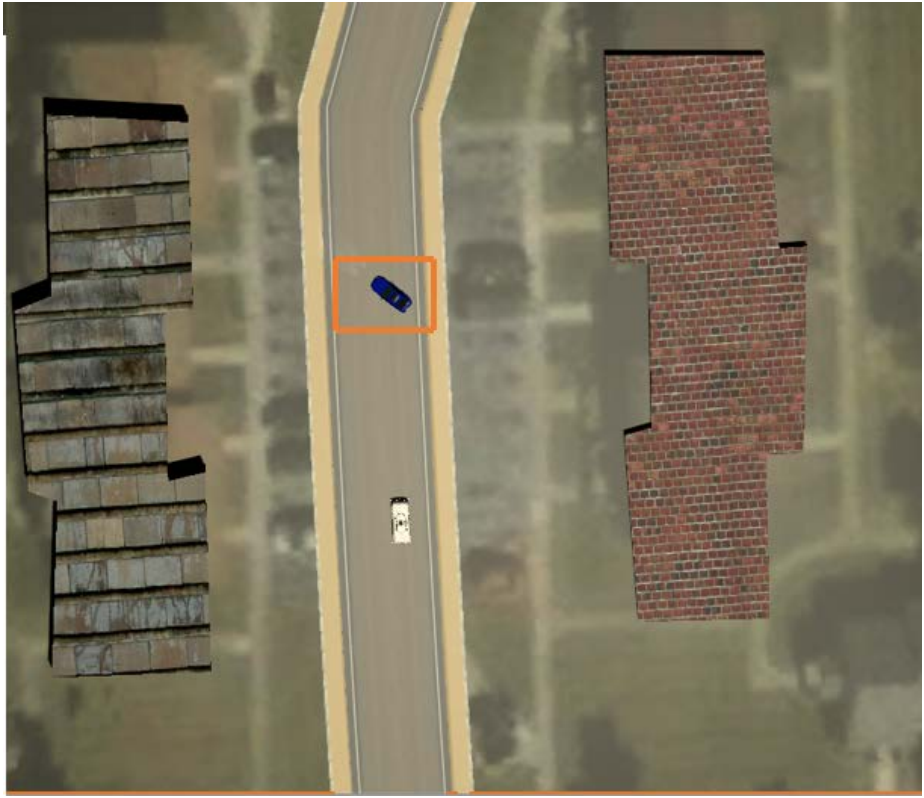

**Figure 48. AV with Sudden Detection of Stopped Obstacle in Front**

Based on those basic scenarios and their extensions in the simulation environment, the safe operation of an AV shuttle can be tested in the Linden Residential Area before actual deployment. Future work will concentrate on making the simulation environment look more realistic, extending the test matrix and preparing an easy to use interface for the evaluation; scalability and portability to other possible AV shuttle deployment sites and AV-in-the-Loop (AViL) evaluation.

# Chapter 7. Conclusions

With the help of AV simulators based on the Unity engine, the Unreal engine or commercially available programs like Carmaker, we were able to simulate different traffic scenarios and events that may take place during the deployment of an AV shuttle along the Linden Residential Area route. In this report, the following topics were discussed, and corresponding results were presented regarding the development and use of the simulation environment and simulations for the Linden AV shuttle:

- *Procedure for creating 3D model of the AV shuttle deployment route*. Three different platforms for modeling the real world data imported from OpenStreetMap were presented.
- *Demonstration of the capabilities of three different simulators comprising of Unity engine based simulation, Unreal engine based simulation and CarMaker*. The advantages of these three simulation approaches was the convenience of environment generation for simulation and the built-in virtual sensors used for synthetic data generation, for example, lidar, camera, radar, GPS and IMU.
- *A basic path following behavior for autonomous shuttle pre-deployment testing* was shown for running the AV shuttle along the route based on NDT matching localization and pure pursuit path following algorithm that were shown in both the Unity engine based LG SVL simulator and the Unreal engine based CARLA simulator.
- *Utilization of the NVIDIA Drive PX2 GPU board* and its related automated driving library, DriveWorks for processing synthetic data within CarMaker simulation, including open space detection, object detection and classification, traffic lane detection etc.

Using the simulators of this report for numerous simulations, several traffic scenarios and events of significance for AV shuttle operation in the Linden Residential Area were identified. Table 7-1 shows these significant scenarios and events that were evaluated in simulation along with simulation improvements being pursued and possible problems that were identified and suggested mitigation strategies.

| Significant Events / Traffic Scenarios | Current Realization in Simulator | Possible Problem and Mitigation Method or Improvement |
|---|---|---|
| AV shuttle path following performance along the shuttle route | Pure pursuit path following algorithm was used initially along the route as it is easy to implement and as the user is expected to replace that with more advanced algorithms. Pure pursuit was initially used as it is very easy to implement. However, when the speed is higher, the deviation from the original path is larger (Figure 34). | Possible Problem: Path following errors may be large at higher speeds and while the AV is following a curved part of the route like the traffic circles at high speed. Improvement: More advanced path following algorithm like the parameter space based robust path following algorithm in [21] will be used in simulation and later AViL testing. Proposed Mitigation: Check AV shuttle company path tracking performance during factory acceptance and Linden area initial testing on both straight and curved parts of the road at operation speed. Use lower speed in curved parts of the road if necessary. |

| Significant Events / Traffic Scenarios | Current Realization in Simulator | Possible Problem and Mitigation Method or Improvement |
|---|---|---|
| Route finish time impacted by different traffic conditions | Add NPC vehicles on the road with different speeds as shown in Figure 49 to determine changes in route finish time. | Proposed Mitigation: Measure the overall route finish time with different traffic density levels and adjust AV shuttle timing and frequency accordingly. Record route travel time during AV shuttle operation for later analysis. |
| Handling non-signalized intersections | Model road intersection in the simulator and add NPC vehicles and pedestrians crossing at the crosswalk as seen in Figure 45 and Figure 47. | Possible Problem: Real vehicles and pedestrians (and bicyclists and scooters) may have unexpected and sudden collision threat generating maneuvers.<br><br>Proposed Mitigation: At non-signalized intersections, slow down and use manual override if necessary. During autonomous operation, follow rules of traffic. Make sure that emergency braking works. Note maximum deceleration during emergency braking. Make sure that operator and all passengers are safely seated and informed that in the rare event that a vulnerable road user or another vehicle creates a sudden collision risk, the AV may exercise emergency braking. |

| Significant Events / Traffic Scenarios | Current Realization in Simulator | Possible Problem and Mitigation Method or Improvement |
|---|---|---|
| Handling traffic circle with stop-controlled entry | The AV comes to a full stop at the stop sign and then waits for other vehicles that may have arrived earlier at the traffic circle and then enters the circle as shown in Figure 50. | Possible Problem: Non-AV traffic may not obey the rules of traffic properly. Some cars may not come to a full stop and wait or they may enter the traffic circle even though they are not supposed to.<br><br>Proposed Mitigation: The AV will always come to a full stop at the stop sign and check all other traffic at the circle and at the other entries. If other vehicles enter the traffic circle even though they have less priority, the AV should always wait. In case of a possible problem, the operator should engage manual override or just pause the AV until the other erratic traffic object has left the traffic circle. All such manual overrides should be recorded. Traffic circle handling should be evaluated during initial testing in Linden. Use of different colored overhead lights (red for wait state, green for operation state for example) or textual information displays can be used to inform other drivers and vulnerable road users of AV intent or next operation mode. |

46 | Smart Columbus Program | Linden Residential Area CEAV Simulation Evaluation – Final Report

| Significant Events / Traffic Scenarios | Current Realization in Simulator | Possible Problem and Mitigation Method or Improvement |
|---|---|---|
| Occlusion for detection at intersections or traffic circle | Simulator has access to ground truth data. Smart intersection and smart traffic circle are also implemented in simulator as shown in Figure 51 to solve this. | Possible Problem: AV sensors may have difficulty in picking up all other traffic including vulnerable road users at non-signalized intersection or traffic circle. This may also be due to occlusions or the AV sensors not covering the whole intersection or traffic circle.<br><br>Proposed Mitigation: Check that the AV has full intersection and traffic circle coverage during Linden area initial tests. Use manual override if necessary. Use low speed at intersections and traffic circles. A longer term proposed mitigation strategy is to use an overhead camera, radar or lidar and a Road Side Unit broadcasting other traffic and vulnerable road user Basic Safety Message (BSM) data to implement a smart intersection and/or traffic circle such that the AV has extra information about the other traffic objects. |
| Weather conditions may impact performance and safe operation | Generate different rendering effect for fog and rain condition in the simulator as shown in Figure 52. Use vehicle dynamics based simulator like Carmaker or add road friction effect to vehicle model to realistically see the effect of weather on AV dynamics. | Possible Problem: Adverse weather like rain, snow, fog and ice will affect performance of perception sensors and also the handling of the AV. Emergency braking during adverse road surface conditions may cause problems.<br><br>Proposed Mitigation: Weather conditions should be recorded during AV operation. The AV should consider suspending service during inclement weather. AV perception sensor operation should be checked during adverse weather by the AV operator. Lower speed of operation is advised during non-ideal weather conditions. |

| Significant Events / Traffic Scenarios | Current Realization in Simulator | Possible Problem and Mitigation Method or Improvement |
|---|---|---|
| Stopped vehicles or other objects on road stalling the AV | The AV stops and waits in the simulator in this case. This will be updated with an automatic, collision free takeover maneuver. | Possible Problem: In the case of stopped traffic like a car with problems that is blocking traffic, current AV shuttles stop and wait. An overtake maneuver should be executed safely.<br><br>Possible Mitigation: The AV shuttle operator should execute a manual overtake maneuver, if possible. |
| Emergency braking need | Emergency braking is used in simple collision avoidance as shown in Figure 53. More advanced collision avoidance method will be added to simulator. | Possible Problem: Sudden emergency braking will be needed as a rare event. Emergency braking response is much faster than the human operator response.<br><br>Proposed Mitigation: Emergency braking should be on at all times. A warning should be issued to the operator and passengers. Emergency braking should be tested during AV shuttle factory testing and initial testing in Linden. |
| Pedestrians | Pedestrians are simulated on both the roadside and on crossroads. The simulated AV uses simple collision avoidance and stops in the presence of a collision risk with pedestrians. | Possible Problem: Emergency braking is usually tuned for other cars which may cause problems when pedestrians (or bicyclists or scooters or pedestrians carrying bicycles) are encountered.<br><br>Proposed Mitigation: AV shuttle emergency braking operation should be thoroughly checked for cars, pedestrians and other vulnerable road users. |

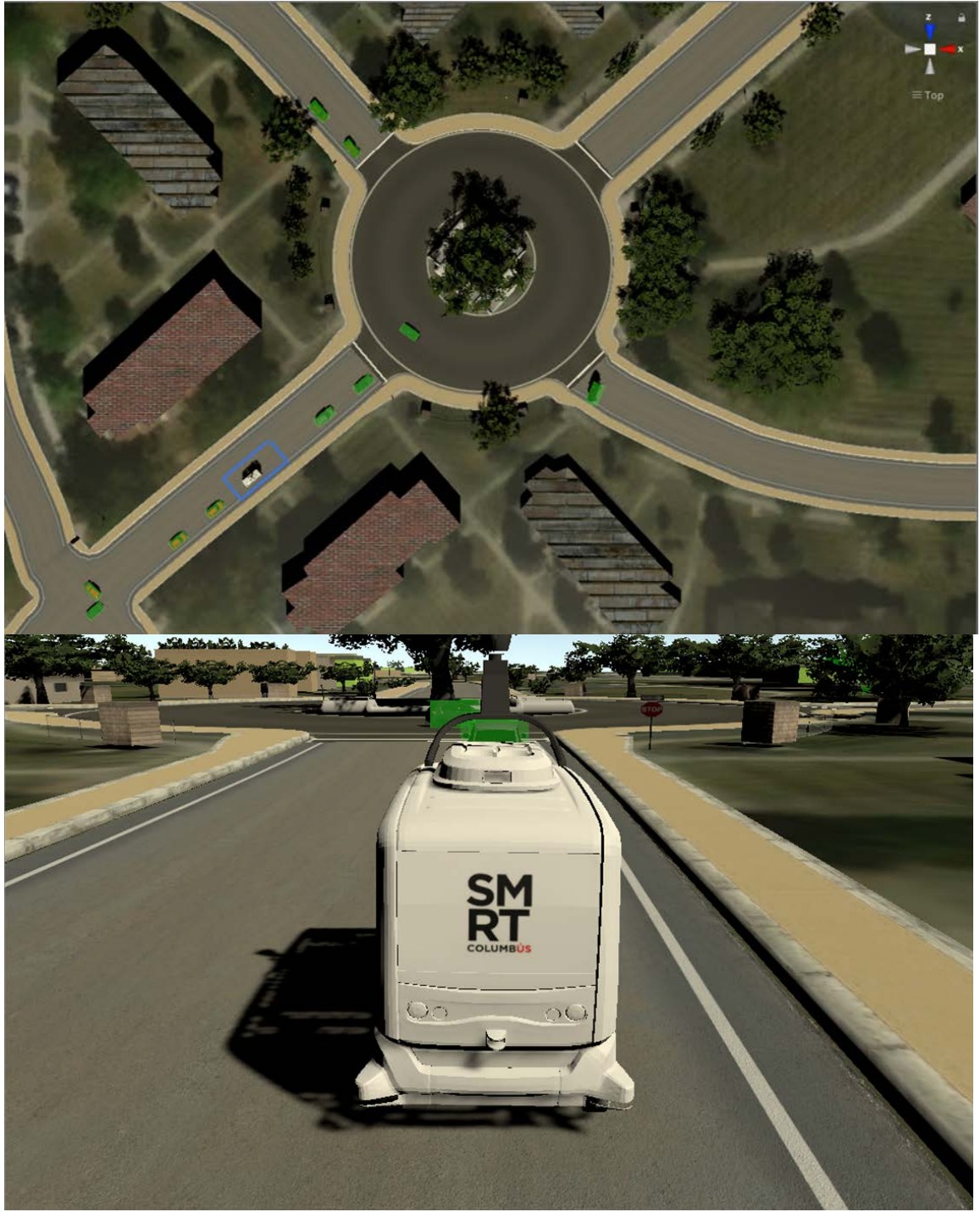

**Figure 49. AV Shuttle in Slow Moving Traffic Fleet. Traversal Time along the Route Is Impacted by Traffic Condition on Road (Ego Vehicle Shown in Blue Box and Other Vehicles in Green Bounding Box, Top View and Driver Camera View)**

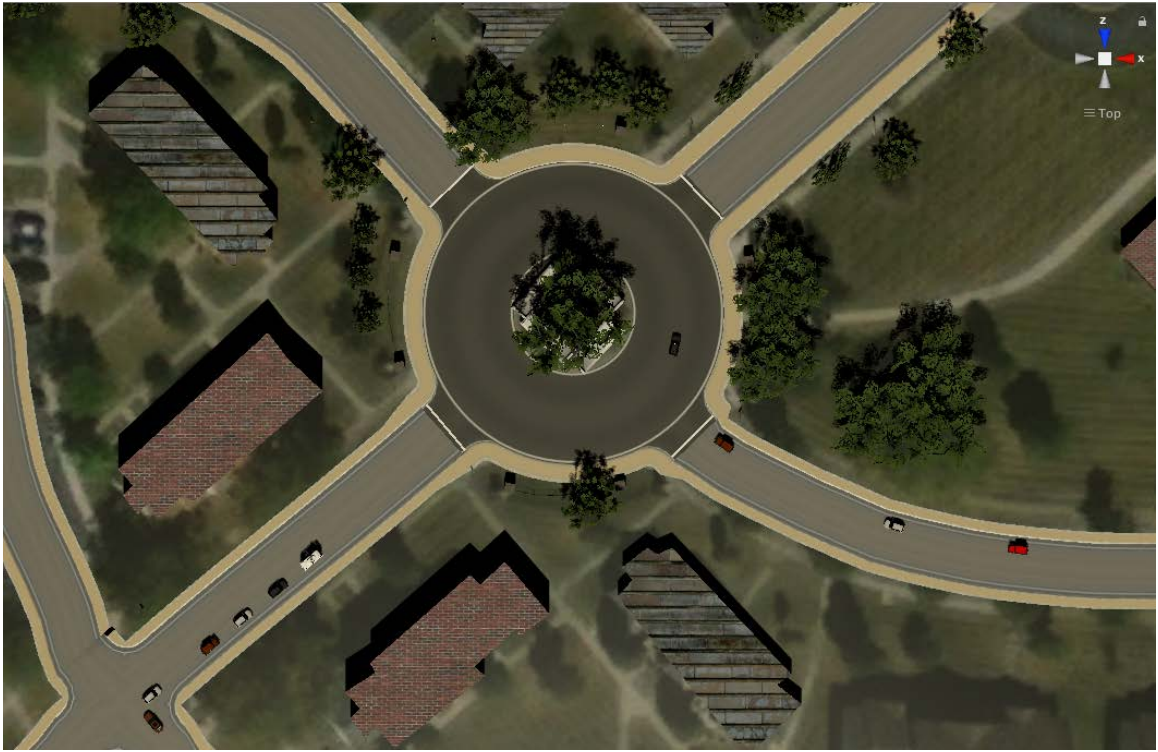

**Figure 50. Realization of AV Shuttle at Traffic Circle. The AV (Black Car on Right Hand Side) Arrives First to Enter the Traffic Circle First. The Next One Will Enter Once the Black AV Exits the Traffic Circle.**

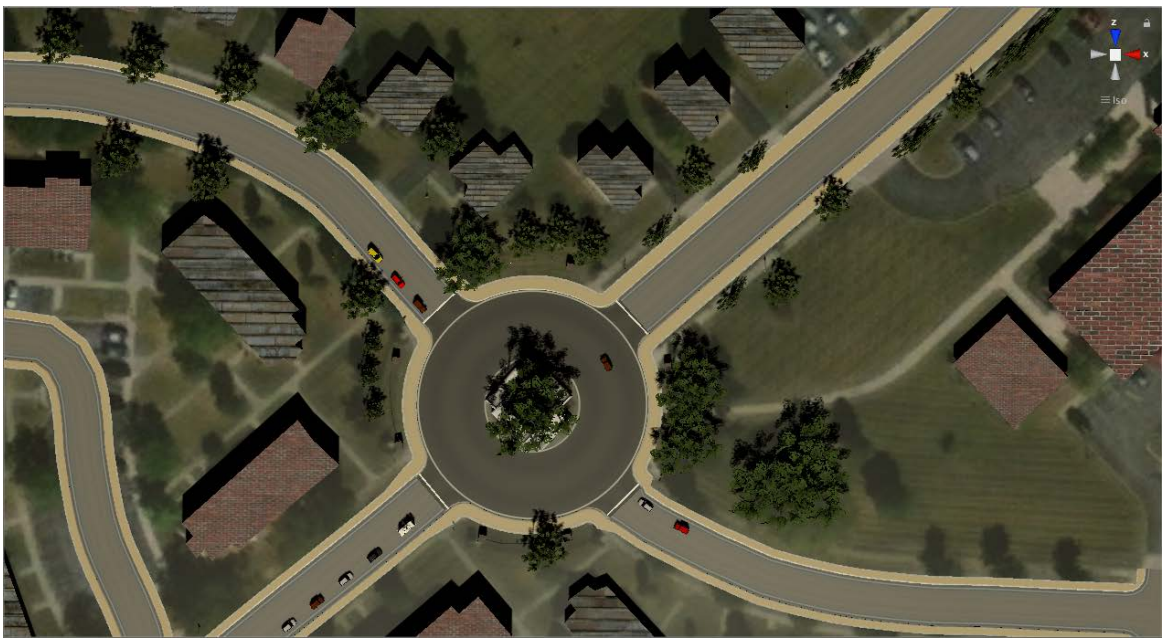

**Figure 51. A Camera Looking Down to the Traffic Circle to Help with Occlusion for Detection. The Camera is stabilized on Top of the Traffic Circle.**

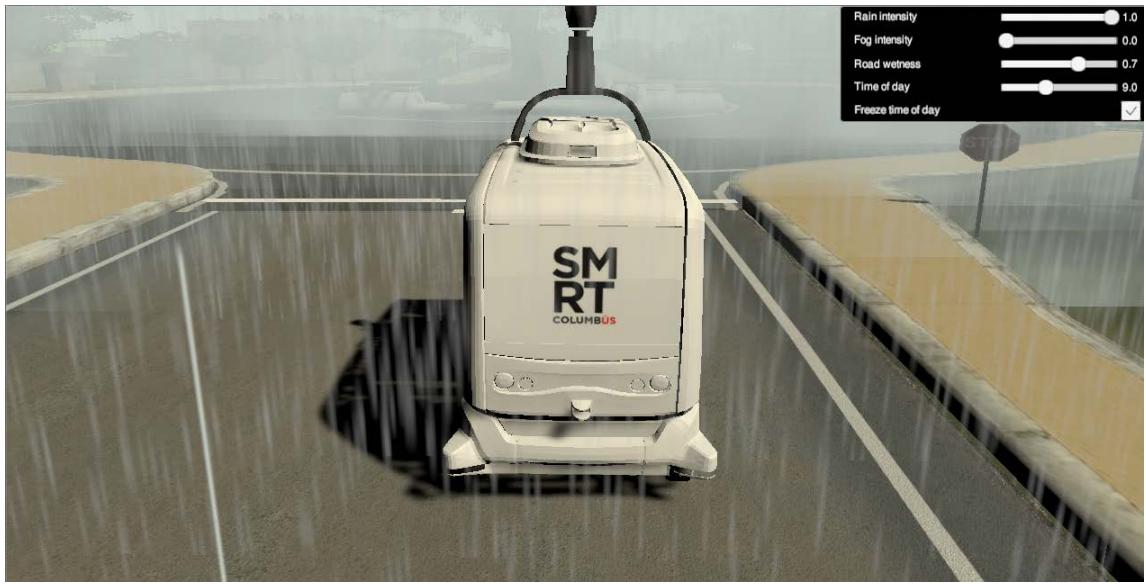

**Figure 52. Weather Condition Realization in the Simulator with Rain**

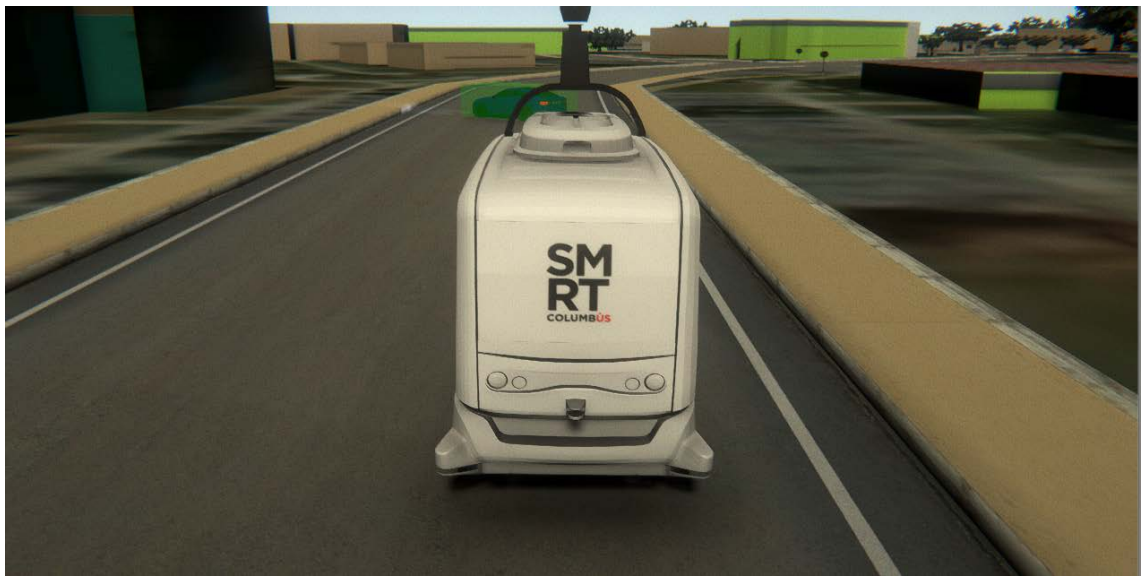

**Figure 53. Current Realization of Stopping Vehicle or Objects. The Vehicle in Green Box is Stopping Vehicle, the AV is Stopping Behind it.**

# Appendix A. Reference List

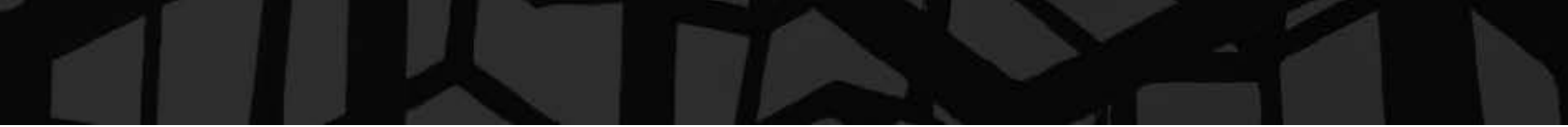

# Appendix B. Acronyms and Definitions

**Table 1: Acronym List** contains project specific acronyms used throughout this document.

**Table 1: Acronym List**

| Abbreviation/Acronym | Definition |
|---|---|
| USDOT | U.S. Department of Transportation |
| SCC | Smart Columbus Challenge |
| AV | Autonomous Vehicle |
| Radar | Radio Detection and Ranging |
| Lidar | Light Imaging, Detection and Ranging |
| IMU | Inertial Measurement Unit |
| GNSS | Global Navigation Satellite System |
| GPS | Global Positioning System |
| ADL | Automated Driving Lab |
| OSU | The Ohio State University |
| RTK | Real Time Kinetic |
| MiL | Model in the Loop |
| HiL | Hardware in the Loop |
| VRU | Vulnerable Road User |
| DSRC | Dedicated Short Range Communications |
| OBU | On Board Unit |
| Fps | Frame Per Second |
| V2V | Vehicle to Vehicle |
| V2I | Vehicle to Infrastructure |
| NTRIP | Network Transport of RTCM via Internet Protocol |
| OSM | Open Street Map |
| CAD | Computer aided design |
| CNN | Convolutional Neural Network |
| RGB | Red/ Green/ Blue system for representing the colors to be used on a computer display |
| PCL | Point Cloud Library |
| 3D | Three Dimension |
| CPU | Central Processing Unit |

**Appendix B. Acronyms and Definitions**

| Abbreviation/Acronym | Definition |
|---|---|
| GPU | Graphical Processing Unit |
| NPC | Non Player Character |
| SLAM | Simultaneous Localization and Mapping |
| ROS | Robot Operating System |
| NDT | Normal Distributions Transform |

*Source: City of Columbus*

# Appendix C. Glossary

**Table 2: Glossary** contains project specific terms used throughout this document.

**Table 2: Glossary**

| Term | Definition |
| --- | --- |
| Autonomous Vehicle | Autonomous vehicle (AV), connected and autonomous vehicle (CAV), driverless car, robocar or robotic car, is a vehicle that is capable of sensing its environment and moving safely with little or no human input. |
| Radar | Radar is a detection system that uses radio waves to determine the range, angle, or velocity of objects. |
| Lidar | Lidar, which stands for Light Imaging Detection and Ranging, is a remote sensing method that uses light in the form of a pulsed laser to measure ranges (variable distances) to the Earth. |
| GNSS | GNSS stands for Global Navigation Satellite System, and is the standard generic term for satellite navigation systems that provide autonomous geo-spatial positioning with global coverage. Global Positioning System (GPS) is a kind of GNSS that is widely used around the world. |
| MiL and HiL Simulation | Simulation is verification steps that take place before the model is deployed into the hardware (autonomous vehicle in our work). Model in the Loop (MiL) simulation is to test the designed controller model. And Hardware in the Loop (HiL) simulation is to test the verified controller with models built in Realtime operating system to simulate real deployment situation. |
| V2V and V2I Communication | In connected vehicle algorithms and functions, vehicle to vehicle (V2V) is communication between autonomous vehicles equipped with DSRC OBU; Vehicle to Infrastructure (V2I) is communication between autonomous vehicle with road side unit. |

*Source: City of Columbus*

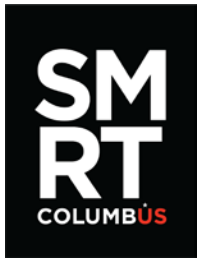